\documentclass[journal]{IEEEtran}
\usepackage{cite}
\usepackage{graphicx}
\usepackage{stfloats}
\usepackage{multirow}
\usepackage{multicol}
\usepackage{diagbox}
\usepackage{float}
\usepackage{enumerate}
\usepackage{pifont}
\newcommand{\cmark}{\ding{51}}%
\newcommand{\xmark}{\ding{55}}%
\usepackage{amsmath,amssymb,amsfonts}
\usepackage{algorithmic}
\usepackage{longtable}
\usepackage{textcomp}
\usepackage{xcolor}
\usepackage{subfig}
\usepackage{threeparttable}
\usepackage[pagebackref=true,breaklinks=true,linkcolor=red,anchorcolor=blue, citecolor=green,colorlinks,bookmarks=true]{hyperref}

\newcommand{\etal}{\textit{et al}. }

\newcommand{\ie}{\textit{i}.\textit{e}., }
\newcommand{\eg}{\textit{e}.\textit{g}., }

\def\BibTeX{{\rm B\kern-.05em{\sc i\kern-.025em b}\kern-.08em
    T\kern-.1667em\lower.7ex\hbox{E}\kern-.125emX}}

\definecolor{DarkBlue}{rgb}{0,0,0}
\begin{document}
\title{Attentional Feature Refinement and Alignment Network for Aircraft Detection in SAR Imagery}
\author{Yan Zhao,
        Lingjun Zhao,
        Zhong Liu,
        Dewen Hu,
        Gangyao Kuang,
        Li Liu
\thanks{Yan Zhao, Lingjun Zhao and Gangyao Kuang are with the State Key Laboratory of Complex Electromagnetic Environment Effects on Electronics and Information System, National University of Defense Technology (NUDT), Changsha, 410073 China \{zy18@nudt.edu.cn, nudtzlj@163.com, kuangyeats@hotmail.com\}.} 
\thanks{Dewen Hu is with the College of Intelligent Science, NUDT, Changsha, China \{dwhu@nudt.edu.cn\}.}
\thanks{Li Liu and Zhong Liu are with the College of System Engineering, NUDT, Changsha, 410073, China. \{dreamliu2010@gmail.com;liuzhong@nudt.edu.cn\}.}
\thanks{This work was supported in part by the National Natural Science Foundation of China under Grant 61872379, 62001480, 62022091, 61825305 and 61806218.}

\thanks{Corresponding author: Lingjun Zhao}
}

\markboth{Submitted to IEEE Transactions on Geoscience and Remote Sensing}%
{Zhao \MakeLowercase{\textit{et al.}}: Aircraft Detection}

\maketitle

\begin{abstract}
Aircraft detection in Synthetic Aperture Radar (SAR) imagery is a challenging task in SAR Automatic Target Recognition (SAR ATR) areas due to aircraft's extremely discrete appearance, obvious intraclass variation, small size and serious background's interference. In this paper, a single-shot detector namely Attentional Feature Refinement and Alignment Network (AFRAN) is proposed for detecting aircraft in SAR images with competitive accuracy and speed. Specifically, three significant components including Attention Feature Fusion Module (AFFM), Deformable Lateral Connection Module (DLCM) and Anchor-guided Detection Module (ADM), are carefully designed in our method for refining and aligning informative characteristics of aircraft. To represent characteristics of aircraft with less interference, low-level textural and high-level semantic features of aircraft are fused and refined in AFFM throughly. The alignment between aircraft's discrete back-scatting points and convolutional sampling spots is promoted in DLCM. Eventually, the locations of aircraft are predicted precisely in ADM based on aligned features revised by refined anchors. To evaluate the performance of our method, a self-built SAR aircraft sliced dataset and a large scene SAR image are collected. Extensive quantitative and qualitative experiments with detailed analysis illustrate the effectiveness of the three proposed components. Furthermore, the topmost detection accuracy and competitive speed are achieved by our method compared with other domain-specific, \textit{e.g.,} DAPN, PADN, and general CNN-based methods, \textit{e.g.,} FPN, Cascade R-CNN, SSD, RefineDet and RPDet.
\end{abstract} 


\IEEEpeerreviewmaketitle

\section{Introduction}
\IEEEPARstart{S}{ynthetic} Aperture Radar (SAR) has attractive imaging capabilities in nearly all weather and illumination conditions and SAR image interpretation has numerous applications in civil and military fields including surface monitoring~\cite{dobson1996knowledge,saha2020building,li2020collaborative}, transportation management~\cite{zhang2020hyperli,biondi2017low}, military surveillance~\cite{cheng2021polsar, dong2019target}, \emph{etc.}. As a fundamental problem in SAR image interpretation fields, target detection aims at accurately locating and identifying targets of interests from SAR imagery and has been researched for several decades~\cite{gao2008adaptive,cui2020ship,zhao2020attention,9351574}. As one of the key problems in SAR target detection, aircraft detection has potential applications in modern airport management, military reconnaissance, \emph{etc.} and has become an independent research direction. Especially with the rapid development of SAR imaging technique, high resolution SAR images can be accessed more easily than before, enabling even more research opportunities for challenging fine-grained SAR target recognition tasks including fine-grained aircraft recognition.



\begin{figure}
\centering
\includegraphics[scale = 0.6]{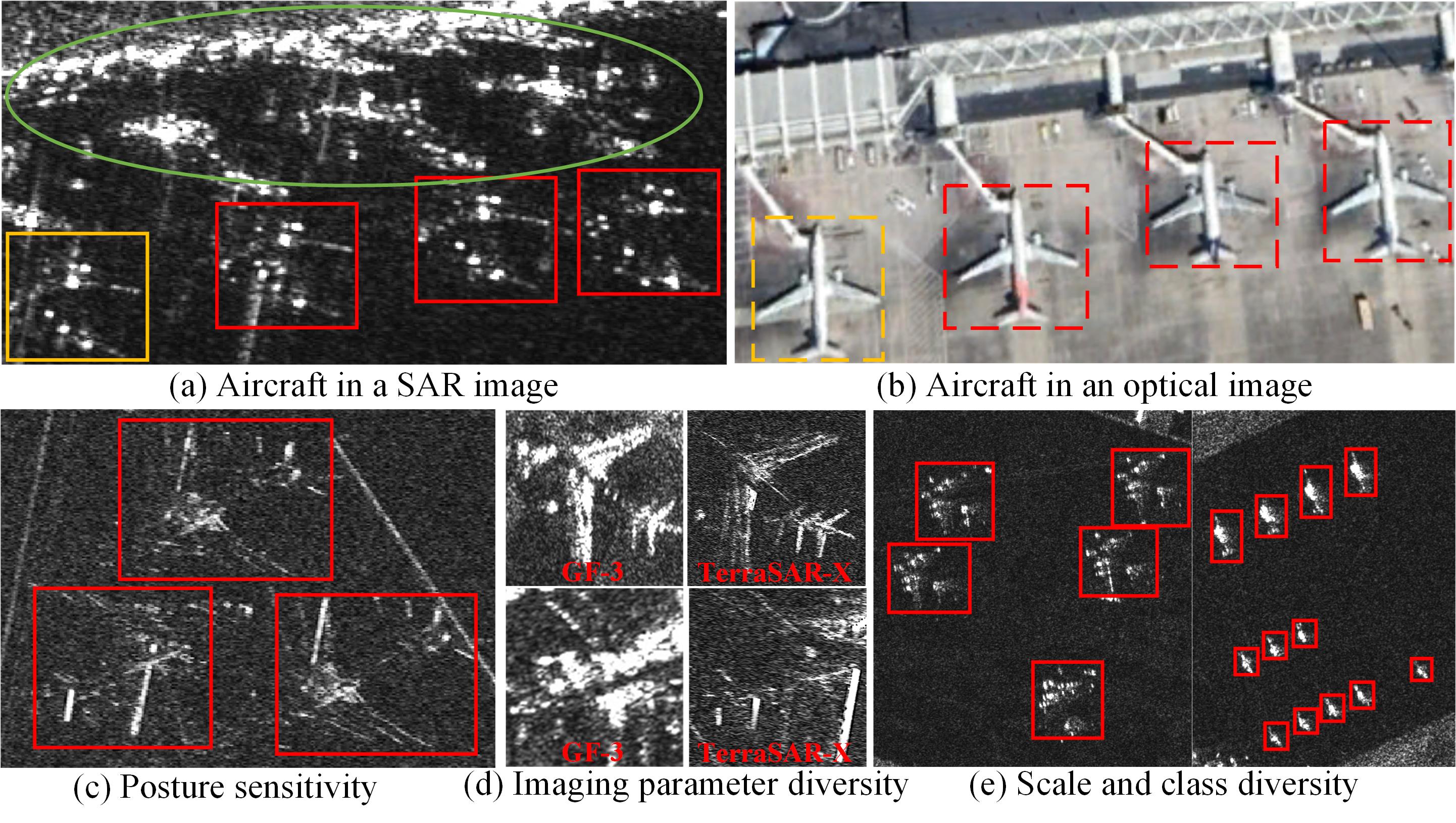}
\caption{\textcolor{DarkBlue}{Aircraft in SAR and optical images.}}  
\label{aircraft_com}
\end{figure}
The problem of SAR aircraft detection has the following main challenges.
\begin{itemize}
\item \textbf{Extremely discrete appearance.} Due to the smooth characteristic of aircraft's surface, the appearance of aircraft in SAR images is discrete compared to those in optical images, leading to the lack of appearance information like geometry and contour cues (see aircraft marked by the real and dotted yellow rectangles in Fig.~\ref{aircraft_com} (a),(b)). Textural information and back-scattering points are important visual cues for SAR aircraft detection.

\item \textbf{Large intraclass variation.} \textcolor{DarkBlue}{Variations in aircraft's posture, scale, category and SAR imaging parameter cause significant impact on the appearance of aircraft in SAR imagery} (\textcolor{DarkBlue}{see Fig.~\ref{aircraft_com} (c) to (e) for examples}), leading to large intraclass variation. Thus, detecting aircraft with large intraclass variation calls for a precise representation of their salient and constant back-scattering information.
%

\item \textbf{Small targets and large scene.} Different from object detection from natural images, \eg ImageNet~\cite{deng2009imagenet} and Microsoft COCO~\cite{lin2014microsoft}, SAR images usually cover very large ranges especially those imaged by satellites and aircraft are with small size and parked sparsely (\textcolor{DarkBlue}{see Fig.~\ref{large_scene_info} for example}), leading to very large searching spaces. Therefore, how to explore context information to locate regions of interest to increase detection speed plays a key role.   

\item \textbf{Uninformative background clutter.} As illustrated by the green ellipse of Fig.~\ref{aircraft_com} (a), complicated background clutters like the scattering clusters of buildings are uninformative and could lead to a number of false alarm aircraft. Therefore, suppressing the disturbance of background clutters is important for accurate aircraft detection. 
\end{itemize} 

Earlier methods for SAR aircraft detection~\cite{dou2016aircraft,zhang2018aircraft,hu2019aircraft} rely on handcrafted gray-scale features and expert interpretation knowledge of aircraft, \eg gradients, structures, contours, and are sensitive to model parameters and lack of generalization to diversified situations, which result in suboptimal detection performance. Recently, Convolutional Neural Networks (CNNs) have achieved remarkable successes for object detection in optical imagery~\cite{liu2019texture,liu2020deep} and also provide great opportunities for aircraft detection in SAR imagery~\cite{dou2016aircraft,wang2017aircraft,guo2018shht,diao2018aircraft, he2018component}.\par 
However, most of current CNN-based aircraft detection methods are directly borrowed from the field of object detection in optical imagery ~\cite{ren2015faster,liu2016ssd,lin2017feature,redmon2017yolo9000}, without fully taking into consideration the aforementioned domain knowledge and challenges of SAR aircraft. That is to say, such detectors are not tailored well for aircraft detection in SAR images, leading to poor interpretability and suboptimal performance.
Specifically, aircraft's low-level textural details are not fully considered as the high-level semantic features. Also, the ability for suppressing interference of surroundings is unsatisfied, which leads to insufficient and suboptimal representation for aircraft in SAR images. Besides, it's inappropriate for capturing aircraft's discrete and irregular back-scattering points by a traditional convolution, of which the convolutional kernels are inflexible. Additionally, detecting sparsely parked small aircraft in large scene images calls for a carefully designed framework with sophisticated detection heads to maintain a trade-off between accuracy and speed, however, is neglected in current methods.  \par 
In response to the aforementioned challenges, a single-shot detector namely Attentional Feature Refinement and Alignment Network (AFRAN) is proposed for detecting aircraft in SAR images with balanced detection accuracy and speed. With RefineDet~\cite{zhang2018single}, \textcolor{DarkBlue}{an excellent CNN-based detector for handling diversified natural objects}, as a basic prototype, \textcolor{DarkBlue}{the novelty of our method stands on a careful consideration of unique characteristics of aircraft, \eg structures, appearance, textures, in SAR images, and is realized by specially designing three crucial modules in AFRAN for feature refinement and alignment.} For feature refinement, Attention Feature Fusion Module (AFFM), which consists of several feature aggregation pathways followed by a Split-Attention (SA) block, is designed for representing both textural and semantic features as well as highlighting significant information of aircraft adaptively. For feature alignment, Deformable Lateral Connection Module (DLCM) applied at lateral connections of a three-layer fine-grained feature pyramid, focus on aligning discrete back-scattering points of aircraft to spots of convolutional kernels with less interference involved. Furthermore, Anchor-guided Detection Module (ADM) is attached at multi-scale feature maps to align sampling points of convolutional kernels with features at unique areas indicated by revised anchors. Our method is evaluated on a self-built SAR aircraft sliced dataset and a large scene image. Extensive quantitative and qualitative experiments illustrate the contributions of the three proposed sub-modules and the comprehensive performance of our method compared with other CNN-based detectors. Additionally, several key factors within our sub-modules are also discussed in detail. \par

The contributions of this paper are summarized as follows: 
\begin{enumerate}[(1)]
    \item A single-shot detector AFRAN is proposed in this paper for aircraft detection in SAR images by carefully taking the characteristics of aircraft into consideration.
    \item To identify and locate aircraft accurately, three significant components, \ie Attention Feature Fusion Module (AFFM), Deformable Lateral Connection Module (DLCM) and Anchor-guided Detection Module (ADM), are designed for refining low-level textural details and high-level semantic features as well as aligning discrete information of aircraft progressively.
    \item The topmost detection accuracy with competitive speed is achieved by our method on a self-built SAR aircraft sliced dataset and a large scene SAR image compared with other CNN-based methods, in which the domain knowledge of aircraft is underutilized.
\end{enumerate}\par 
The remainders of this paper are organized as follows. In Section II, the related works for aircraft detection in SAR images are provided. In Section III, motivations and our method are introduced in detail. Experiments and analysis are given in Section IV. Finally, Section V concludes the work of this article. 
 
\section{Related Works}
\begin{figure}[H]
\centering
\includegraphics[scale = 0.62]{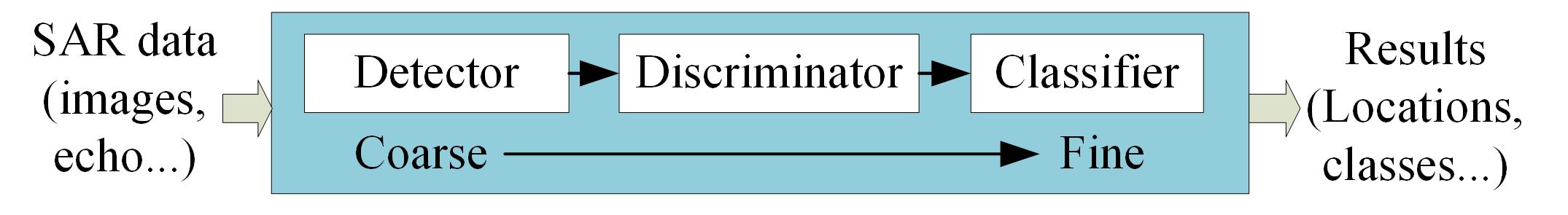}
\caption{The processing flow of SAR ATR.} 
\label{SAR_ATR}
\end{figure}

\textit{\textbf{SAR Automatic Target Recognition (SAR ATR).}} 
SAR Automatic Target Recognition (SAR ATR) generally refers to detect and recognize target signatures from SAR data~\cite{ross1999sar,novak2000state,novak1997automatic}. A standard architecture of SAR ATR proposed by MIT Lincoln Laboratory~\cite{dudgeon1993overview} is provided in Fig.~\ref{SAR_ATR}, which contains three progressive stages including detection, discrimination and classification. Among the three steps, target detection acting as a footstone of the downstream tasks, aims at coarsely figuring out potential regions of interests (targets) and eliminate background areas. A discriminator is adopted to identify non-objects and objects from outputs of the detector. Finally, a classifier is constructed to identify the classes of targets by using diversified domain-specific knowledge~\cite{zhao2001support,chen2016target,olson1997automatic}. 
\par 
\textit{\textbf{Aircraft detection in SAR images.}}
Currently, methods for aircraft detection in SAR images could be coarsely divided into knowledge-based and CNN-based paradigms. \par
\textbf{Knowledge-based methods.} In these ways, apron areas are firstly located by statistic models based on gray-scale features of backgrounds. Then, potential aircraft within these regions are identified by employing various hand-crafted features according to prior knowledge of aircraft, \eg shapes, contours, attributes of back-scattering points. Among these methods, Hu \etal~\cite{hu2019aircraft} introduced a Generalized Gamma Mixture Distribution (GGMD) based detector to detect aircraft in non-homogeneous backgrounds. Dou \etal~\cite{dou2016aircraft} proposed a similarity measurement to identify candidates by using Kullback–Leibler Divergence (KLD) of Gaussian Mixture Model (GMM) based on saliency maps and scattering structure features of aircraft. Besides, Zhang \etal~\cite{zhang2018aircraft} introduced an active shape model (ASM) for contour evaluation by utilizing geometric information of aircraft. Needless to say, aircraft could be detected by designing complicated statistic models and combining various knowledge of aircraft together. However, the detection performance and generalization of these methods usually degenerates drastically where obvious mismatches exist between inflexible hand-crafted features and diversified appearance of aircraft. \par 
\textbf{CNN-based methods.} The CNN-based object detectors could be finely divided into two mainstream paradigms including two-stage and one-stage detection frameworks. Inspired by R-CNN~\cite{girshick2014rich} and Fast R-CNN~\cite{girshick2015fast}, Ren \etal~\cite{ren2015faster} proposed Faster R-CNN, which laid down an original end-to-end two-stage processing framework. And the detection accuracy has been improved continually by numerous successors~\cite{lin2017feature,cai2018cascade,pang2019libra}. In contrast of improving detection accuracy regardless of time consumption, A considerable detection speed without much performance degradation is pursued by one-stage detectors~\cite{liu2016ssd,redmon2016you,redmon2018yolov3,tian2019fcos}.\par 
With respect to aircraft detection in SAR images by using CNN-based methods, Wang \etal~\cite{wang2017aircraft} and Guo \etal~\cite{guo2018shht} employed a CNN-based classifier to identify aircraft's candidates within suspicious areas. To regress locations of aircraft accurately, Diao \etal~\cite{diao2018aircraft} employed a Fast R-CNN to discriminate aircraft. Considering partial characteristic of aircraft, He \etal~\cite{he2018component} introduced a component-based detector based on YOLOv2~\cite{redmon2017yolo9000}, which consists of paralleled root and part detectors, to identify and match fuselages and wings of aircraft. Besides, Guo \etal~\cite{guo2020scattering} proposed an attention pyramid network based on Feature Pyramid Network to explore scattering information enhancement (SIE) of aircraft. To handle sparse and discrete scattering points of aircraft, Zhao \etal~\cite{9046763} introduced a Pyramid Attention Dilated Network (PADN) based on RetinaNet~\cite{lin2017focal} by designing a Multi-Branch Dilated Convolutional Module (MBDCM). Also, an Attention Feature Fusion Network (AFFN)~\cite{affn} is designed for aircraft detection in SAR images. Undoubtedly, aircraft could be detected by these methods stably benefiting from powerful feature representation ability of CNN. However, the lacks of fully considerations of aircraft's domain-knowledge and the challenges limit a further promotion boosted by these methods.

\section{Methodology}
In this section, key inspirations of our method for tackling the aforementioned problems are described in subsection $A$. And the holistic architecture of our method and its inner modules including AFFM, DLCM and ADM, are depicted in subsections $C$, $D$ and $E$, respectively. Loss functions are provided at the end of this section. 
\subsection{Motivations}
Feature refinement and alignment are main concentrations of our method for tackling domain-knowledge and challenges of aircraft detection in SAR images. \par 
\textbf{Feature refinement.} A balanced representation of aircraft's low-level textural details and high-level semantic features with powerful suppression of interference is essential for detecting aircraft effectively. Although detailed information of small aircraft could be extracted by a low-level feature map, the semantic information it contains is rare and the correlations among different aircraft's features are weak. A wider receptive field could be obtained by a deeper neural network. However, textural details of aircraft may be indistinguishable due to a long propagation pathway, in which several intermediate convolutions involved. Instead of employing additional bottom-up pathways~\cite{liu2018path,tan2020efficientdet}, combining low-level textural details and high-level features directly could acquire multi-level aircraft's features distinctly without losing much original textural information. Additionally, serious interference caused by complex surroundings should also be resolved. Although significant features could be refined and highlighted \cite{hu2018squeeze,woo2018cbam,park2018bam}, a layer-aware attention mechanism should be specially cared considering the differences of aircraft's characteristics across multi-level feature channels.\par 
\textbf{Feature alignment.} A highly constant alignment between discrete back-scattering features and convolutional spots is essential for locating and identifying aircraft precisely. In comparison to traditional convolution, of which convolutional kernels' sampling locations are fixed, \eg $3\times3$, deformable convolution~\cite{dai2017deformable,zhu2019deformable} could extract features within irregular areas by using convolutional kernels, of which sampling locations are tuned by trainable offsets. Hence, it's naturally suitable for representing discrete deformable features of aircraft accurately with the alleviation of deformable convolution in our method. Besides, a progressive detector striking the trade-off between accuracy and speed is practical and essential for detecting sparsely parked small aircraft in SAR images.\par
Shortly, the low-level textural and high-level features of aircraft are fused and emphasized throughly by feature refinement. Also, the significant discrete features of aircraft are captured precisely by feature alignment. With the aforementioned two criteria as guidelines, our method could perform competitively.

\subsection{Overall Architecture of AFRAN}
\begin{figure}
\centering
\includegraphics[scale = 0.5]{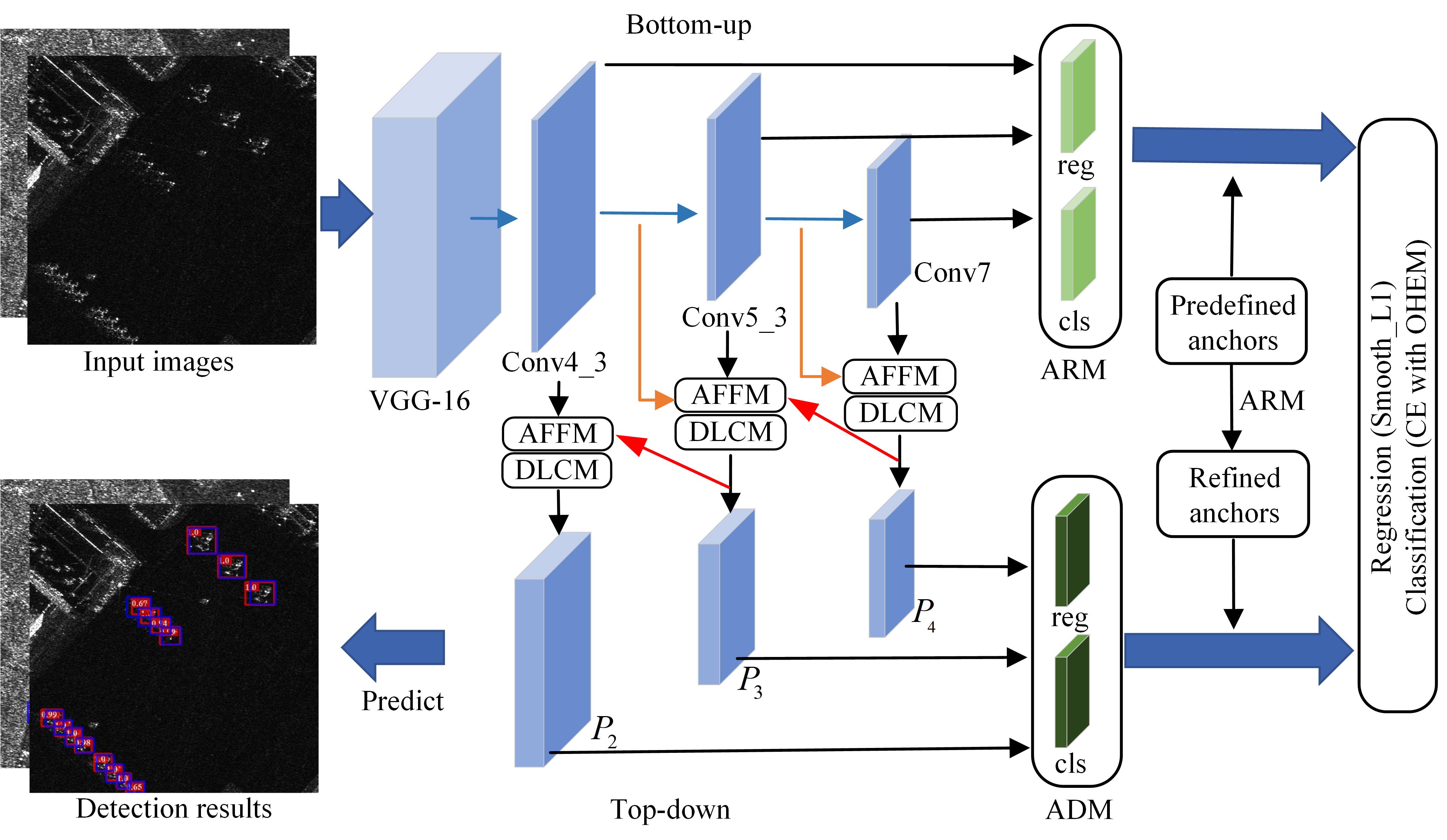}
\caption{The architecture of AFRAN.} 
\label{AFRAN}
\end{figure}

Our method belongs to a one-stage detector, of which the architecture is illustrated in Fig.~\ref{AFRAN}. \par 
\textbf{Bottom-up pathway.} A truncated VGG-16 network~\cite{simonyan2014very} acted as a backbone, is employed to extract fundamental features from the input SAR images in the bottom-up pathway. Three intermediate layers of VGG-16 including Conv4\_3, Conv5\_3 and Conv7, are selected as hierarchically semantic features of aircraft. The spatial sizes of the three feature maps are 8, 16 and 32 times down-sampling of the original input images, which are $80\times80$, $40\times40$ and $20\times20$ in pixels, respectively. \par 
\textbf{Top-down pathway.} A fine-grained feature pyramid is established by combining and refining the three-layer features selectively in the top-down pathway. Three groups of AFFM are firstly adopted to fuse and filter semantic information from the three-layer feature maps. Here two feature forward propagation pathways (orange arrows in Fig.~\ref{AFRAN}) are introduced to fully utilize low-level textural details of aircraft. Considering much interference around aircraft may be extracted by traditional convolution with axis-aligned kernels, three groups of DLCM are introduced following AFFM to capture aircraft's discrete characteristics adaptively. Most specifically, $P_4$ is set up at the first group of AFFM and DLCM by leveraging Conv\_7 and Conv5\_3. $P_3$ is constructed by the second group of AFFM and DLCM by using Conv5\_3, Conv4\_3 and $P_4$ together. And the low-level fine-grained feature map $P_2$ is established by fusing Conv4\_3 and $P_3$ at the third group of AFFM and DLCM.\par 
\textbf{Coarse-to-fine detection head.} Identifying aircraft progressively could alleviate mismatches between initial anchors and ground truths on location and quantity. Drawing from a coarse-to-fine detection strategy\cite{zhang2018single}, an Anchor Refinement Module (ARM) and an improved object detection module namely Anchor-guided Detection Module (ADM)~\cite{zhang2020refinedet++} are introduced to our method for predicting aircraft progressively. Specifically, refined anchors are firstly acquired by adjusting initial anchors based on parameterized locations and confidences, which are predicted by feeding Conv4\_3, Conv5\_3 and Conv7 to ARM. Aircraft are predicted eventually by feeding features, where locations at $P_2$, $P_3$ and $P_4$ are calculated by refine anchors, to ADM subsequently. \par 
To optimize our method, a multi-task loss including binary and multi-class cross entropy losses with Online Hard Example Mining (OHEM)~\cite{shrivastava2016training} and Smooth L1 loss, is employed for supervising classification and regression branches of ARM and ADM, simultaneously.

\subsection{Attention Feature Fusion Module}
\begin{figure}[H]
\centering
\includegraphics[scale = 0.76]{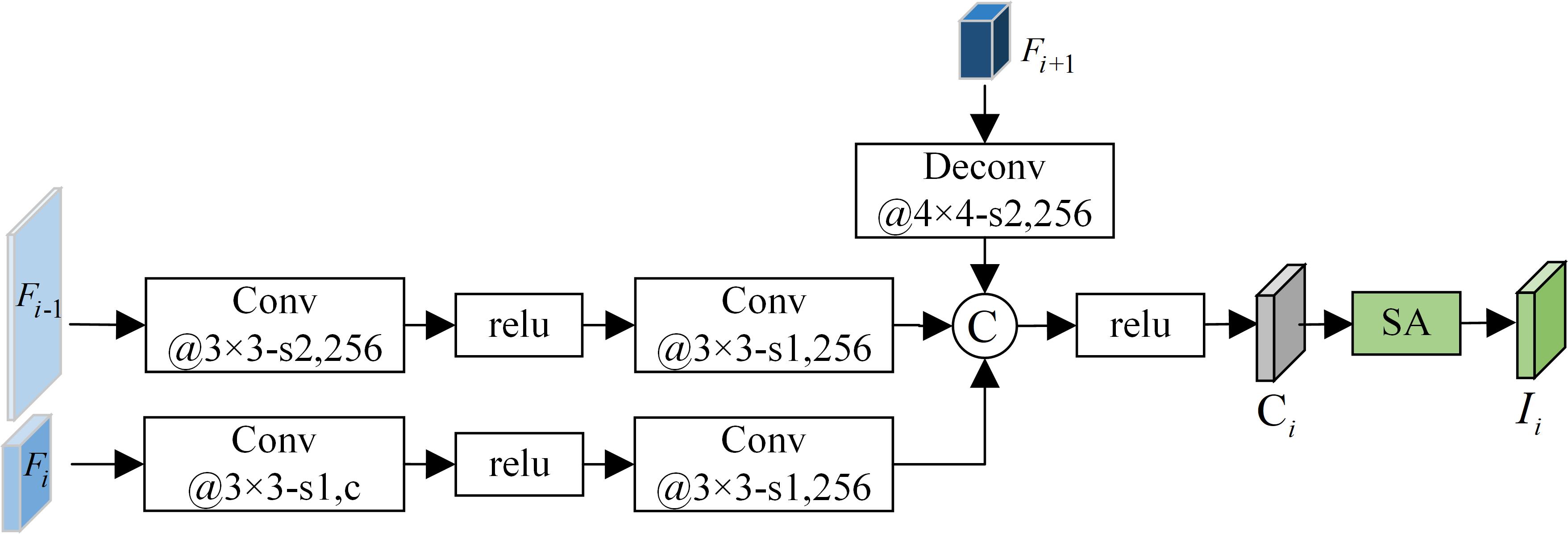}
\caption{Structure of AFFM.} 
\label{AFFM}
\end{figure}

To leverage aircraft's low-level textural details and semantic features harmoniously, an Attention Feature Fusion Module (AFFM) as illustrated in Fig.~\ref{AFFM}, is introduced when building the fine-grained feature pyramid of AFRAN.\par 
 \par 
\textbf{Multi-level feature aggregation.} To acquire multi-dimension features, a $3\times3$ convolution with stride of 2 followed by a relu activation function is firstly applied at the ${i-1}$ th basic feature map $F_{i-1}$ to reduce its spatial size for feature fusion.\textcolor{DarkBlue}{Meanwhile, the ${i+1}$ th basic feature layer $F_{i+1}$ is up-sampled by a $4\times4$ deconvolution to acquire abundant and discriminative semantic features, which are helpful for detecting small aircraft from low-level feature layers. To satisfy requirements of different modules, one and two layers of $3\times3$ convolution layers with a stride of 1 are attached at the $i-1$ th feature map $F_{i-1}$ and the $i$ th feature map $F_{i}$, respectively, to maintain discrepancy as well as relaxing interference of different feature stages.} After that, the modified $F_{i+1}$, $F_{i-1}$ and $F_{i}$ with the same spatial sizes are obtained. After concatenating (\textcircled{c}) the three intermediate features along channel dimension, an intermediate feature map $C_i$ containing abundant textural and semantic features of aircraft, is obtained. Specifically, Conv5\_3 and Conv7 are adopted to build the topmost intermediate feature $C_4$. The middle intermediate feature $C_3$ is constructed by leveraging Conv4\_3, Conv5\_3 and $P_4$. The low-level intermediate feature map $C_2$ is obtained by combining Conv4\_3 with $P_3$ finally. \par 

\begin{figure}
\centering
\includegraphics[scale = 1.35]{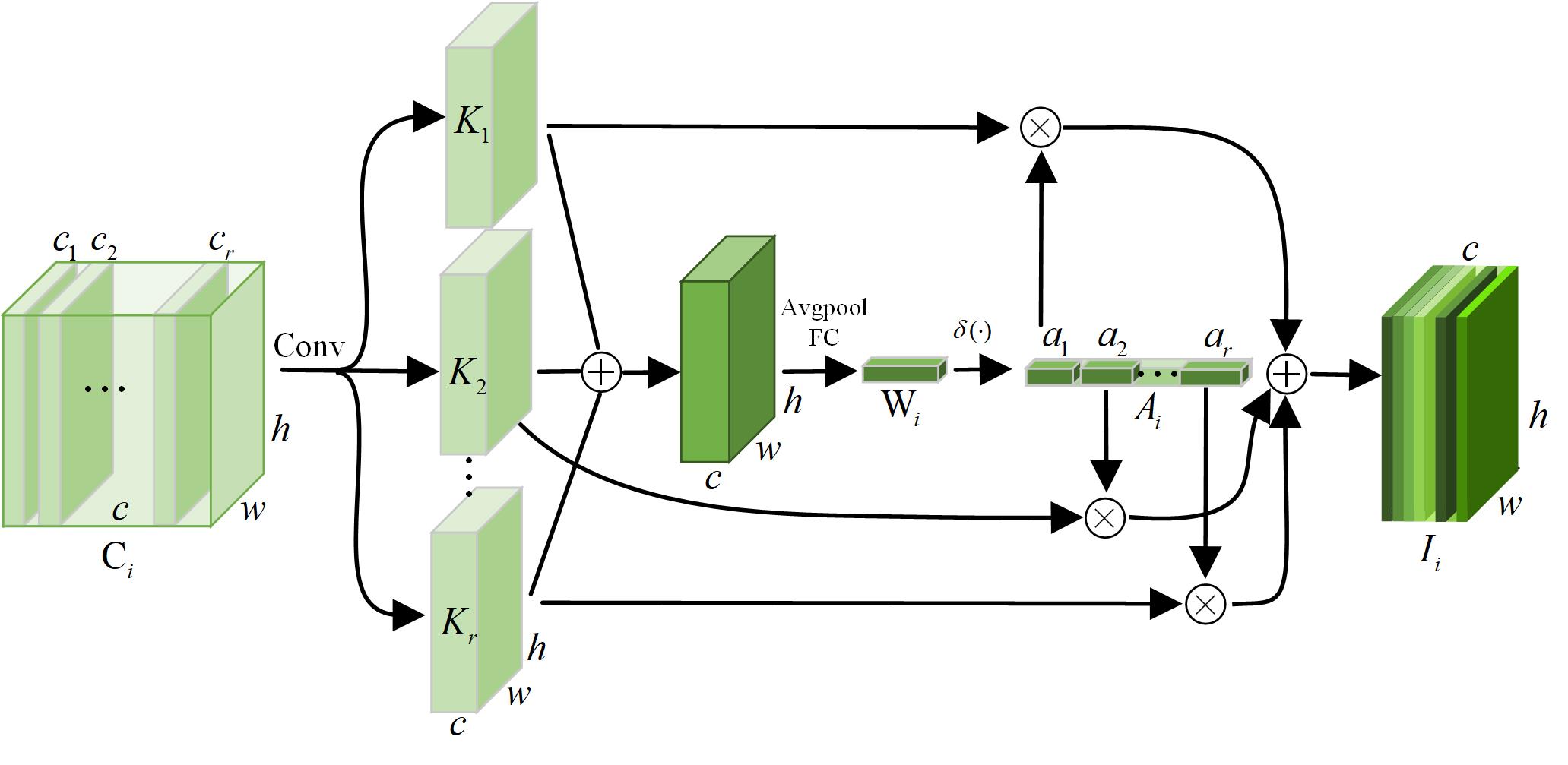}
\caption{Structure of SA.} 
\label{SAB}
\end{figure}

\textbf{Feature refinement by Split-Attention.} Although abundant features of aircraft are accumulated after concatenation, gaps across different feature layers are still obvious. Also, interference of uninformative background around aircraft may also be collected. To highlight significant features of aircraft and suppress background interference at different feature layers carefully, a layer-aware attention mechanism namely Split-Attention (SA)~\cite{zhang2020resnest} is employed following the concatenated feature $C_i$. The structure of SA is illustrated as Fig.~\ref{SAB}.
 \par 
Let $\mathbf{C}_{i}\in \mathbf{R}^{h\times w\times c}$ denotes a concatenated feature block of $r$ feature maps, where $h$, $w$, $c$ and $r$ refer to block's height, width, channel number and the number of concatenated features within AFFM, respectively in our method. To acquire refined feature block $I_i$, $r$ groups of individual feature maps ($\textbf{\textit{K}}_{1}$, $\textbf{\textit{K}}_{2}$, ... $\textbf{\textit{K}}_{r}$), of which numbers of channels all equal to $c$, are firstly generated by sending $\mathbf{C}_{i}$ to a convolution (Conv) with ${c}\times{r}$ filters and splitting along channel dimension. Then, these $r$ groups of features are added element-wisely ($\oplus$) and sent into an adaptive two-dimension average pooling (Avgpool) and fully connection (FC) in sequence to produce a weighted vector $\mathbf{\textbf{\text{W}}}_{i}$. It represents contributions of different channels of the concatenated feature map. Then, $r$ groups of channel attention weights $a_{1}$, $a_{2}$, ... $a_{r}$ derived from $\mathbf{A}_{i}$, is acquired by applying a Softmax function ($\sigma$) on $\mathbf{\textbf{\text{W}}}_{i}$. Finally, the layer-aware refined feature map ${\mathbf{I}_{i}}\in {\mathbf{R}^{h\times w\times c}}$ could be acquired by an element-wise addition ($\oplus$) of $r$ groups of productions ($\otimes$) between $\mathbf{\textit{K}}_{r}$ and the corresponding vector $a_{r}$.  
\subsection{Deformable Lateral Connection Module}
\begin{figure}[H]
\centering
\includegraphics[scale = 1.5]{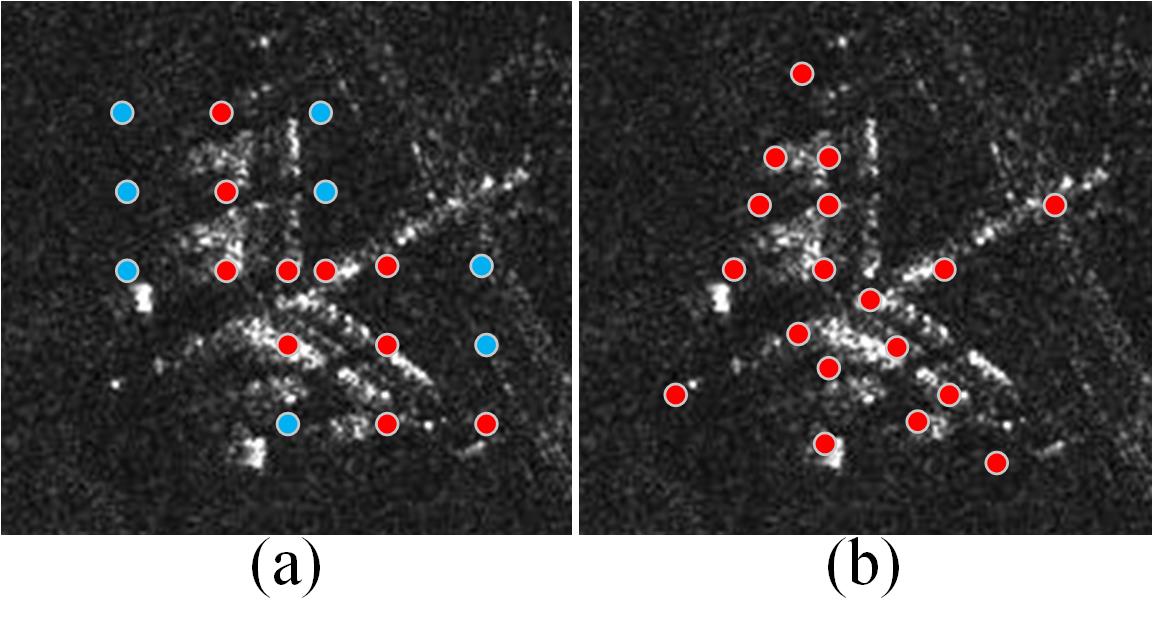}
\caption{Sampling locations of convolutional kernels for aircraft and surroundings by different types of convolution.(a) locations sampled by vanilla convolutional kernels. (b) locations sampled by deformable convolutional kernels.} 
\label{dcn_sample}
\end{figure}

\begin{figure}[H]
\centering
\includegraphics[scale = 0.55]{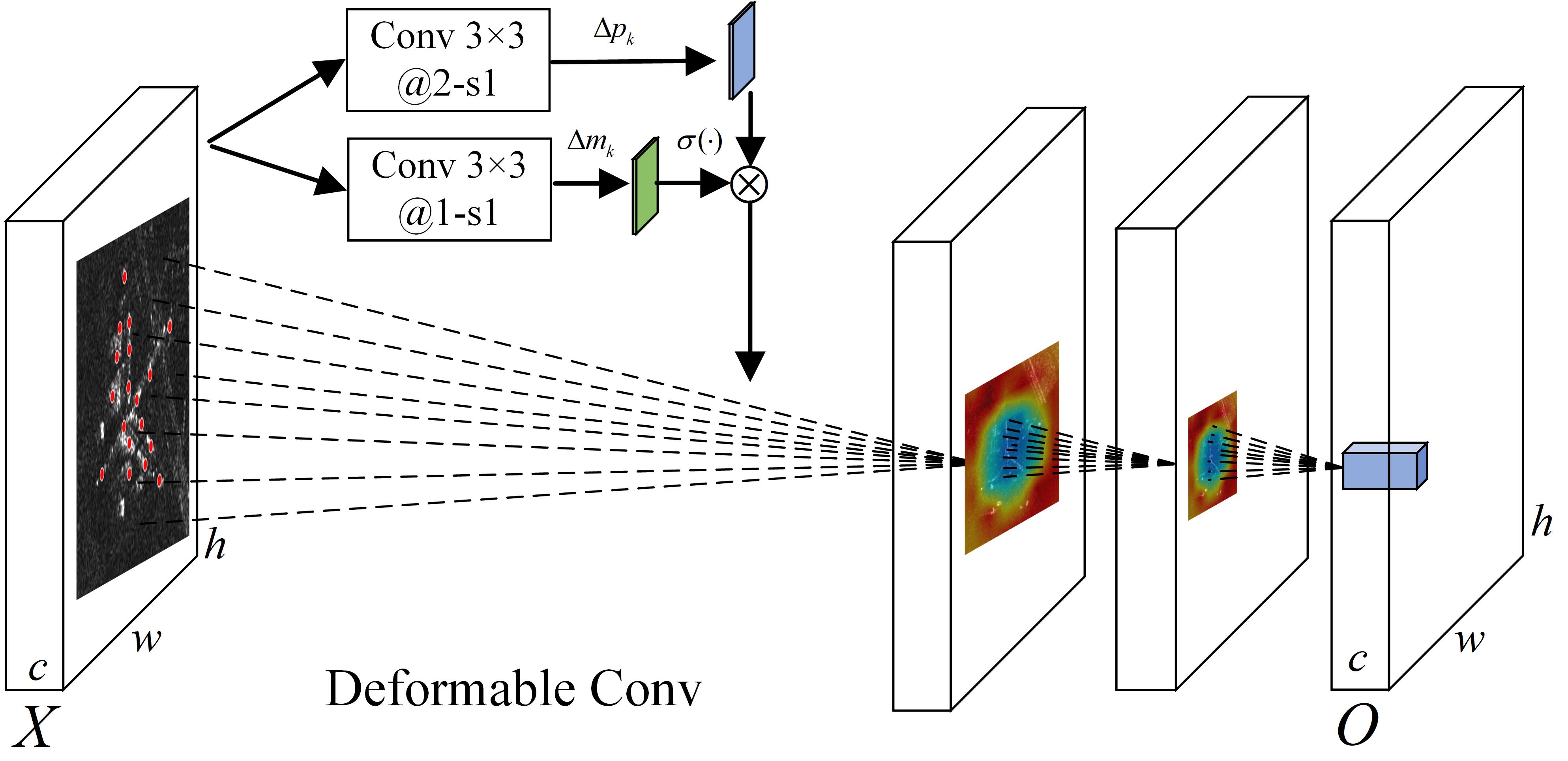}
\caption{Structure of DLCM.} 
\label{DLCM}
\end{figure}

Representing discrete yet constant features of aircraft by a suitable convolution is essential for accurate aircraft detection in SAR images. As a toy example depicted in Fig.~\ref{dcn_sample}, red and blue points refer to sampling locations of convolutional kernels for aircraft and surroundings, respectively.
As shown in Fig.~\ref{dcn_sample} (a), back-scattering information of surroundings around aircraft is easily captured by a traditional convolutional operation due to its regular and rigid sampling strategies. However, as shown in Fig.~\ref{dcn_sample} (b), a deformable convolution~\cite{dai2017deformable,zhu2019deformable} is naturally appropriate for capturing discrete and significant features of aircraft without introducing much background interference benefiting from its deformable convolutional strategy. Based on the superiority, three groups of Deformable Lateral Connection Module (DLCM) with several deformable convolution stacked, as depicted in Fig.~\ref{DLCM}, are designed in our method for capturing aircraft's discrete features following AFFM.\par
As illustrated by the blue and green rectangles of Fig.~\ref{DLCM}, two vanilla $3\times3$ convolution with stride 1 outputs 2 and 1 are firstly attached on the input feature map $\mathbf{\textit{X}}$ to generate a two-dimension offset map $\Delta p_{k}$ and a score mask $\Delta m_{k}$, respectively. Therefore, revised sampling locations of the deformable convolution could be acquired by adjusting original axis-aligned convolutional spots through $\Delta p_{k}$. And  discrete deformable features of aircraft are generated after modulating deformable values with the score mask $\Delta m_{k}$. In our method, flexible and irregular information of aircraft could be represented effectively by stacking several deformable convolution in DLCM sequentially. \par
\par 
Concisely, the deformable convolution is formulated in Eq.~(\ref{dcn}). Here $\mathbf{X}(p_n+p_k+\Delta p_k)$ refers to deformable input features. They are acquired by adjusting original eight-neighborhood features $\mathbf{X}(p_n+p_k)$ of $\mathbf{X}(p_n)$ using offsets $\Delta p_k$, where $k = (i,j)$ $ i,j \in -1, 0, 1$ for a $3\times3$ convolutional kernel. The final output matrix $\mathbf{Y}(p_n)$ is obtained after mapping and modulating $\mathbf{X}(p_n+p_k+\Delta p_k)$ through trained weights $\mathbf{W}_k$ and modulation factor $\Delta m_{k}$ simultaneously.
\begin{equation}
\label{dcn}
\mathbf{Y({{p}_{n}})=\sum\limits_{k}^{K}{\mathbf{W}_{k}}\cdot \mathbf{X}({{p}_{n}}+{{p}_{k}}+\Delta {{p}_{k}})}\cdot \Delta {{m}_{k}}
\end{equation} 

\subsection{Anchor-guided Detection Module} 
Due to the lack of well-designed feature cropping strategies, \eg ROI Align~\cite{he2017mask}, ROI Wrapping~\cite{ren2015faster} leveraged in two-stage detectors, one-to-many mappings and obvious misalignments between anchors and features in one-stage detectors easily incur inaccurate identification and location for small aircraft. Instead of representing aircraft by a set of representative points~\cite{yang2019reppoints} or features aligned by initial anchors~\cite{chen2019revisiting}, an Anchor-guided Detection Module (ADM) derived from~\cite{zhang2020refinedet++}, is adopted in our method for establishing a strict single-mapping between refined anchors and features. Most specifically, unique features calculated by refined anchors are utilized for identifying aircraft ultimately.\par
\begin{figure}[H]
\centering
\includegraphics[scale = 1.0]{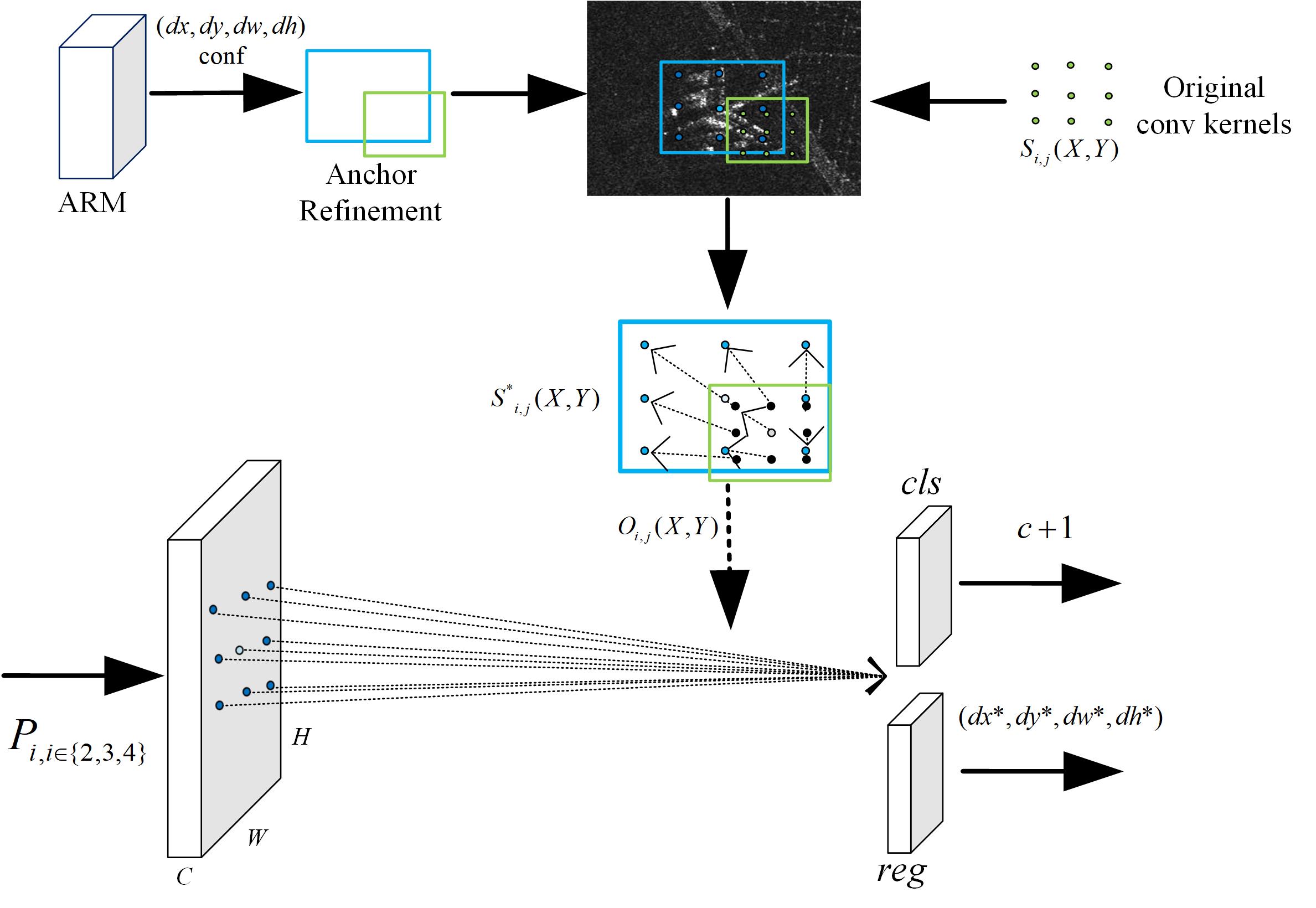}
\caption{Structure of ADM.} 
\label{ADM}
\end{figure}

 
As illustrated in Fig.~\ref{ADM}, blue rectangle refers to a refined anchor generated by adjusting locations and suppressing redundancy of initial anchors based on parameterized offsets ($dx, dy, dw$,$dh$) and confidence produced by ARM. It distinct that features covered by a refined anchor are more representative than those covered by a initial anchor (the green rectangle in Fig.~\ref{ADM}) resulting from an improved overlap between the revised anchor and the ground truth aircraft. Thus, a higher correspondence between features captured by the revised sampling points $S^{*}_{i,j}(X,Y)$ (9 blue points in Fig.~\ref{ADM}) and the aircraft could be established than before. Based on advantages of deformable convolution, two-dimension offsets $O_{i,j}(X,Y)$ between revised sampling points ($S^{*}_{i,j}(X,Y)$) and the related original sampling points ($S_{i,j}(X,Y)$) could be calculated and acted as deformable convolutional offsets for final regression and classification. \par 
\par 
Concisely, the processing flows of ADM are formulated as Eq.~(\ref{aadm_basic}),~(\ref{aadm_refine}) and (\ref{aadm_offsets}). The original sampling points $S_{i,j}(X,Y)$ are formulated as Eq.~(\ref{aadm_basic}), where $i$, $j$ $\in \{0,1,k-1\}$ for a $3\times3$ convolution.
\begin{gather}
\begin{split}
&S_{i, j}(X,Y)=(X-\left\lfloor\frac{k}{2}\right\rfloor+i+0.5, Y-\left\lfloor\frac{k}{2}\right\rfloor+j+0.5)
\label{aadm_basic}
\end{split}
\end{gather}\par
The corresponding aligned sampling points $S^*_{i, j}(X,Y)$ at feature maps $\textbf{F}$ are formulated as Eq.~(\ref{aadm_refine}), where $x1, y1, x2, y2$ refer to the locations of refined anchors.
\begin{equation}
\begin{split}
& S^{*}_{i, j}(X,Y) = \\ & (\frac{k x_{1}+\left(x_{2}-x_{1}\right)(i+0.5)}{k S},
\frac{k y_{1}+\left(y_{2}-y_{1}\right)(j+0.5)}{k S} ),\\
&  i,j\in {0,1,2,3...k-1}
\end{split}
\label{aadm_refine}
\end{equation} \par 
The two-dimension offsets for a deformable convolution formulated as Eq.~(\ref{aadm_offsets}), are obtained by subtracting Eq.~(\ref{aadm_refine}) from Eq.~(\ref{aadm_basic}).
\begin{equation}
\begin{split}
O_{i}(X)   & = S^{*}_{i, j}(X)- S_{i, j}(X) \\
           & = \frac{x_{1}}{S}-X+\left\lfloor\frac{k}{2}\right\rfloor+\left(\frac{x_{2}-x_{1}}{ S}-1\right)(i+0.5)  \\
O_{j}(Y)   & = S^{*}_{i, j}(Y)- S_{i, j}(Y)  \\
           & = \frac{y_{1}}{S}-Y+\left\lfloor\frac{k}{2}\right\rfloor+\left(\frac{y_{2}-y_{1}}{w S}-1\right)(j+0.5)  
\end{split}
\label{aadm_offsets}
\end{equation}\par 
\subsection{Loss Functions}
The overall loss for optimizing our network provided by Eq.~(\ref{Eq_total}), contains losses of ARM and ADM. As defined by Eq.~(\ref{Eq_sigle}), each part of Eq.~(\ref{Eq_total}) is a weighted sum of the classification loss~($L_{conf}$) between predicted classes ($x$) and truth labels ($c$) and the regression loss~($L_{reg}$) among predicted classes ($x$), bounding boxes ($l$) and ground truth locations ($g$).
\begin{equation}
\label{Eq_total}
    {{L}_{total}}={{L}_{ARM}}+{{L}_{ADM}}
\end{equation}
\begin{equation}
\label{Eq_sigle}
    L(x,c,l,g)=\frac{1}{N}\left( {{L}_{conf}}(x,c)+\alpha {{L}_{reg}}(x,l,g) \right)
\end{equation}\par
Where $N$ refers to the number of positive anchors matched to any ground truth objects. As defined by Eq.~(\ref{Eq_conf}), $L_{conf}$ is a binary or multi-class cross entropy loss between the predicted classes $x$ and ground truth labels $c$. 
\begin{equation}
\label{Eq_conf}
L_{conf}(x, c)=-\sum_{i \in \text {Pos}}^{N} x_{i,j}^{p} \log \left(\hat{c}_{i}^{p}\right)
\end{equation}\par
The Iverson bracket indicator function $x_{i,j}^{p}$ is 1 where the $i$ th anchor matches to $j$ th ground truth object for class $p$ otherwise 0. $\hat{c}_{i}^{p}$ refers to a Softmax loss defined as Eq.~(\ref{Eq_softmax}).
\begin{equation}
\label{Eq_softmax}
\hat{c}_{i}^{p}=\frac{\exp \left(c_{i}^{p}\right)}{\sum_{p} \exp \left(c_{i}^{p}\right)}
\end{equation}\par
For regression, $L_{reg}$ formulated as Eq.~(\ref{Eq_smooth}), refers to a Smooth L1 loss of offsets between predicted bounding boxes ($l$) and ground truths ($g$), where $c_x$,~$c_y$,~$w$ and $h$ are centers, widths and heights of initial or refined anchors and matched ground truths. \par
\begin{equation}
\label{Eq_smooth}
L_{reg}(x,l,g)=\sum_{i \in \text {Pos}}^{N} \sum_{m \in\{c_x, c_y, w, h\}} x_{i j}^{k} \operatorname{Smooth}_{\mathrm{L} 1}\left(l_{i}^{m}-\hat{g}_{j}^{m}\right)
\end{equation}\par
In training phase, the initial anchors are firstly revised by predictions of ARM. Only refined anchors with higher confidence than a preset threshold $\theta$ will contribute loss of ADM. 

\section{Experiments and Analysis}
\subsection{Datasets and Experimental Setup}
\textbf{Datasets descriptions.} Since there is no publicly available dataset for aircraft detection in SAR images, a self-built aircraft sliced dataset and a large scene SAR image are collected for investigating the detection performance of our method in the experiments. \par 
\begin{enumerate}[a)] 
\item \textbf{Aircraft sliced dataset.} \textcolor{DarkBlue}{A self-built aircraft sliced dataset is constructed by using 174 large scene SAR images collected from Chinese GF-3 and German TerraSAR-X satellites. The large scene images captured by GF-3 satellite working in C-band HH and VV polarization and SpotLight (SL) observing mode are with nominal resolution of 1.0 meter. Another images captured by TerraSAR-X satellite working in X band and SpotLight (SL) mode are with nominal resolution of 0.5 meter. The ground truth aircraft were manually annotated by experts of SAR ATR by considering both prior knowledge and the corresponding optical images. After random cropping, 2317 non-overlapped $640\times 640$ slices are collected with 6781 aircraft, of which structures, outlines and main components are clear and wings ranging from about 15 meters to 75 meters in total}. The distributions of bounding boxes' sizes in pixels and aspect ratios are given by Fig.~\ref{dataset_info} (a), (b), respectively. \textcolor{DarkBlue}{Also, some slices of our dataset are provided as Fig.~\ref{Dataset} (a) and (b), in which real aircraft encompassed by blue rectangles are imaged by GF-3 and TerraSAR-X satellites, respectively.} 
\begin{figure}[H]
\centering
\includegraphics[scale = 0.30]{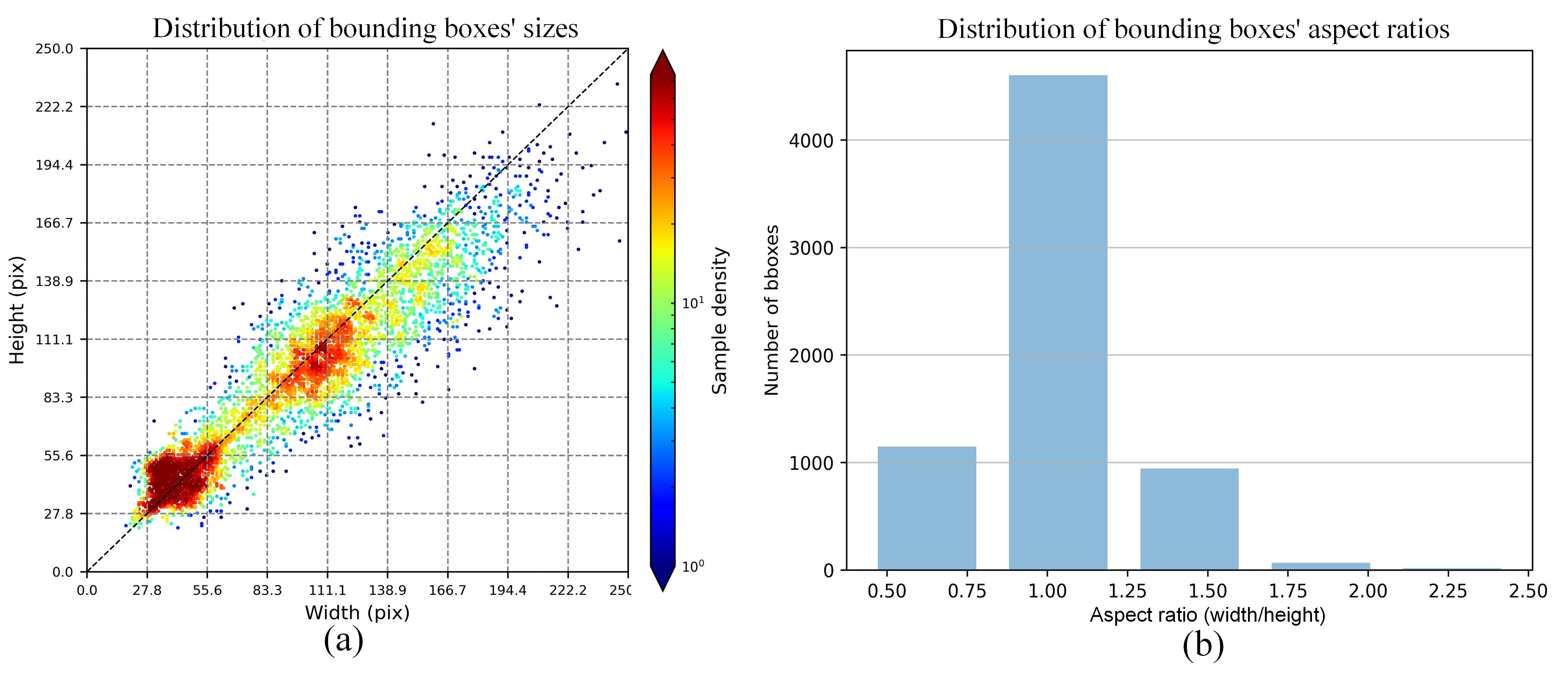}
\caption{Distributions of bounding boxes' sizes and aspect ratios of the self-built aircraft sliced dataset.} 
\label{dataset_info}
\end{figure}

\begin{figure}[H]
\centering
\includegraphics[scale = 0.52]{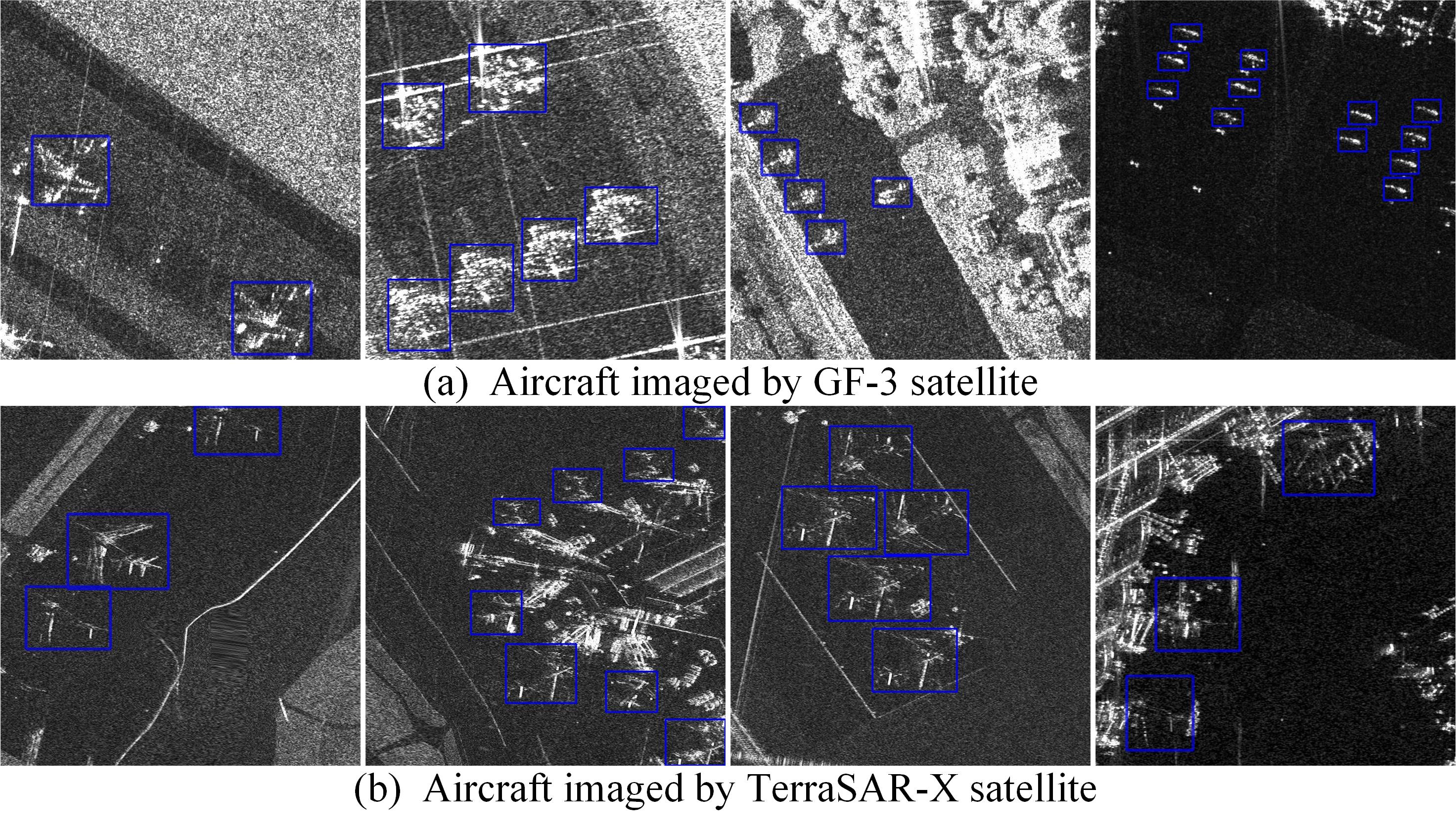}
\caption{\textcolor{DarkBlue}{Slices of SAR images in the self-built aircraft sliced dataset.}}
\label{Dataset}
\end{figure}

\item \textbf{Large scene SAR image.} A large scene image is also adopted for evaluating the performance of our method. The left and the right subfigures of Fig.~\ref{large_scene_info} illustrate the large scene image including an airport and corresponding optical image cropped from Google Earth software, respectively. Some imaging parameters are listed concisely in Table~\ref{large_scene_params}

\begin{figure}[H]
\centering
\includegraphics[scale = 0.52]{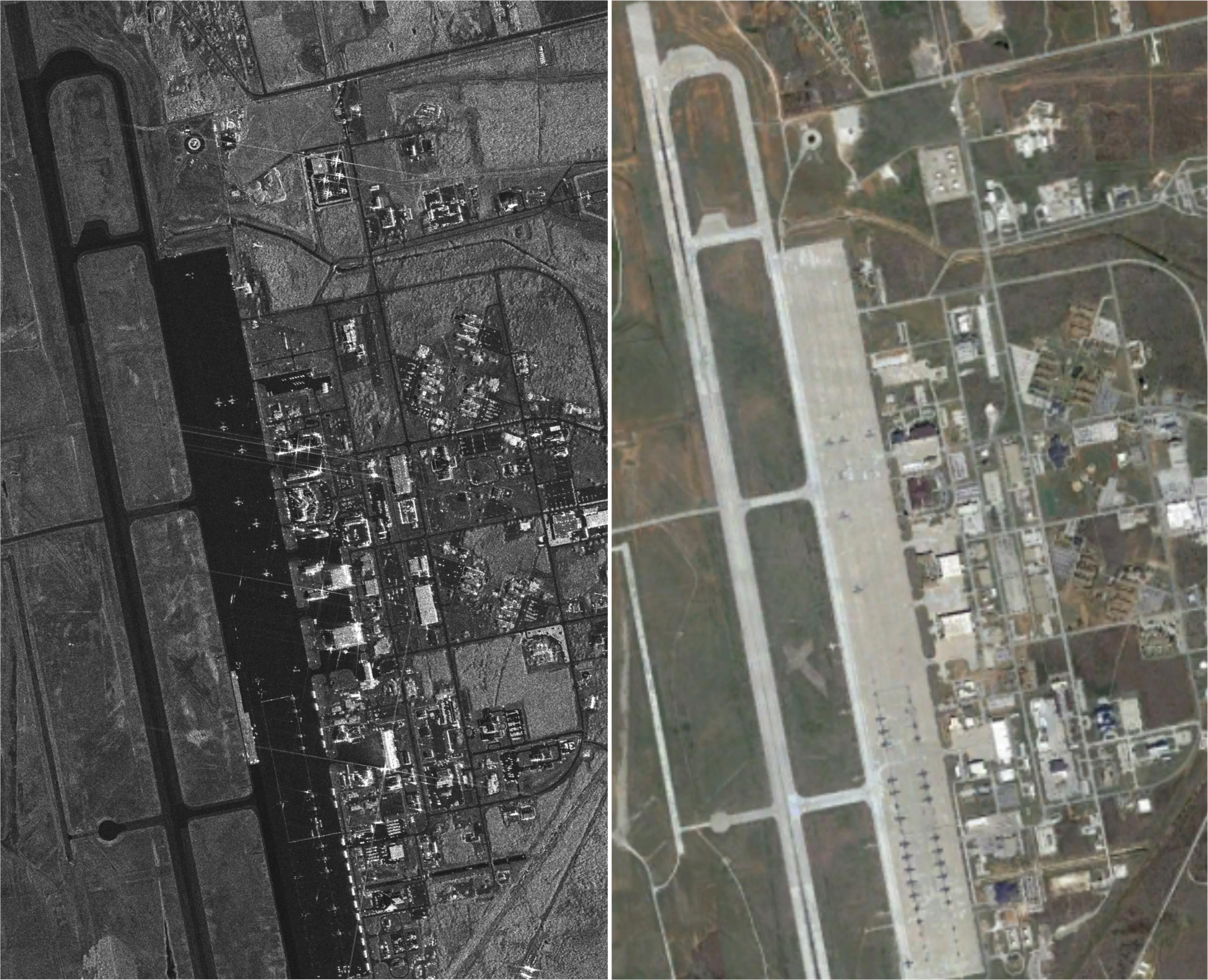}
\caption{A large scene for aircraft detection.}
\label{large_scene_info}
\end{figure}

\begin{table}[H]
\caption{\label{large_scene_params} {\textcolor{DarkBlue}{Imaging parameters of the large scene image}}}
\centering 
\def\arraystretch{1.0} 
\begin{tabular}{c|c}
\hline
Parameter name   & Value \\
\hline
\hline
Product level           & Level 2              \\
Product serial number   & 3253042              \\
Product resolution      & 1.0 (meter)            \\
Imaging central location & (99.82525 W, 32.412828 E) \\
Product height and width & $25784 \times 23161$ (pixel) \\
\hline
\end{tabular} 
\end{table}
\end{enumerate}

\textbf{Implementation details.} The training, valid and test sets are constructed by dividing the original slices with a ratio of 5:2:3. Thus, numbers of images in the three sets are 1158, 463, and 696, respectively. Data augmentation strategies including contrast, illumination distortion, mirroring, random flipping, expanding and cropping, are adopted to increase diversity of aircraft in training set. To maintain considerable overlaps between initial anchors and aircraft, the basic anchor scales for $P_2$, $P_3$ and $P_4$ of our method are set to 32, 64 and 128, respectively. Also, three aspect ratios \{0.5,1.0,2.0\} are assigned to each anchor for all pyramid levels. Our method is trained for 200 epochs with a minibatch of 4. The initial learning rate is 1e-3 and decayed at 75 and 150 epochs with rate ($\gamma$) 0.1. To achieve a stable convergence, warm-up is enabled at the first 5 epochs. Stochastic Gradient Descent (SGD) with weight decay rate 5e-4 and momentum 0.9 is adopted to optimize network parameters. Other compared methods are implemented based on mmdetection framework~\cite{chen2019mmdetection}. 

\textbf{Evaluation metrics.} In all experiments, six average precision indicators from Microsoft COCO~\cite{lin2014microsoft} are adopted including $AP$, $AP^{\mathrm{.5}}$,$AP^{\mathrm{.75}}$,$AP^{\mathrm{s}}$,$AP^{\mathrm{m}}$ and $AP^{\mathrm{l}}$, to judge performance of different methods. Moreover, $AP^{\mathrm{.5}}$ and $AP^{\mathrm{.75}}$ evaluate average precision scores at 0.5 and 0.75 Intersection of Union (IoU) thresholds between predictions and ground truths, respectively. $AP^{\mathrm{s}}$,$AP^{\mathrm{m}}$ and $AP^{\mathrm{l}}$ refer to average precision scores of methods for detecting small, middle and large aircraft averaged by ten IoU thresholds (0.5, 0.55,..., 0.95). Additionally, precision (P), recall (R) and F score ($F_1$) are also adopted. The six AP metrics are calculated at 0.05 confidence threshold. Precision, recall and $F_1$ are acquired at 0.5 confidence threshold. All these metrics are acquired at 0.5 IoU threshold. In terms of time and space complexities, frame-per-second (FPS), model parameter volumes (Params) and multiply-accumulate operations (MAC) are employed, of which the last two metrics are defined as Eq.~(\ref{params}) and (\ref{mac}), respectively. 

\begin{equation}
 \text{Params}  = {{C}_{out}}\cdot ({{k}_{w}}\cdot {{k}_{h}}\cdot {{C}_{in}}+1)
\label{params}
\end{equation}

 $C_{out}$, $C_{in}$, $k_{w}$ and $k_{h}$ in Eq.~(\ref{params}) are the output, input channels and kernel sizes of a convolution. 
\begin{equation}
\text{MAC} = {{C}_{out}}\cdot {{C}_{in}}\cdot {{k}_{w}}\cdot {{k}_{h}}\cdot H_{o}\cdot W_{o}
\label{mac}
\end{equation}

Here $k_{w}$, $k_{h}$, $C_{out}$ and $C_{in}$ are width, height, output and input channels of a convolution. $H_{o}, W_{o}$ are spatial sizes of an output feature map.\par 
Additionally, precision-recall curves of different AP metrics and visual detection results are also provided for further judgements.

\subsection{Ablation Studies}
\subsubsection{\textbf{\textit{Quantitative analysis}}}
The quantitative contributions of different inner modules are evaluated by enabling them progressively and the results are shown in Table~\ref{effect_modules}. A baseline detector is constructed by removing the three modules from AFRAN but maintaining the basic architecture, \ie the three-layer fine-grained feature pyramid. Besides, we disentangle AFFM into feature fusion especially feature forward (orange arrows in Fig.~\ref{AFRAN})  and Split-Attention block, donated as FF and SA, and explore their contributions to aircraft's textural feature extraction and significant feature refinement in detail. Considering the similar architecture and processing scheme between RefineDet and  our method. The detection results of RefineDet are also provided at the first column of Table~\ref{effect_modules}. According to Table~\ref{effect_modules}, several insightful findings could be summarized as follows.\par 

\begin{table*}[!htbp]
\centering
\begin{threeparttable}
\caption{\label{effect_modules}Effects of Inner Modules.}
\def\arraystretch{1.0} 
\begin{tabular}{c|cccc|ccc|cccccc|c}
\hline
Models & FF & SA & DLCM & ADM & P & R & $F_1$  & $AP$ & $AP^{.5}$ & $ AP^{.75}$ & $ AP^{s}$ & $ AP^{m}$& $ AP^{l}$ & FPS \\ 
\hline
\hline
 RefineDet & -  & - & - & - & 0.815 & 0.935 & 0.871 & 0.530 & 0.932 & 0.547 & 0.388 & 0.521 & 0.549  & 60 \\
\hline
\multirow{4}{*}{Baseline\tnote{*}}
& -  & - & - & - & 0.829 & 0.933 & 0.878 & 0.520 & 0.933 & 0.535 & 0.385 & 0.515 & 0.536 &  \textbf{63} \\
 & \cmark & - & - & - & 0.827 & 0.926 & 0.874 & 0.522 & 0.932 & 0.517 & 0.371 & 0.508 & 0.543 & 58 \\
 & \cmark & \cmark & - & - & 0.871 & 0.931 & 0.900 &  0.536 & 0.941 & 0.563 & 0.427 & 0.524 & 0.555 &  50 \\
 & \cmark & \cmark & \cmark & - & 0.899 & 0.927 & 0.913  & 0.538 & 0.929 & 0.572 & 0.394 & 0.526 & 0.564 &  46\\
 \hline
  AFRAN (Ours) & \cmark & \cmark & \cmark & \cmark  & \textbf{0.904} & \textbf{0.932} & \textbf{0.918} & \textbf{0.554} & \textbf{0.941} & \textbf{0.597} & \textbf{0.481} & \textbf{0.537} & \textbf{0.576} & 40 \\
 \hline
 \end{tabular}
 \begin{tablenotes}
    \footnotesize
    \item[*] The baseline detector is a modified AFRAN, which only contains three layers of multi-scale features for prediction. By enabling different sub-modules progressively, performance for detecting aircraft in SAR images is improved. Bold texts indicates the best value in each row.  
  \end{tablenotes}
\end{threeparttable}
\end{table*}
\begin{figure*}
\centering 
\includegraphics[scale = 0.4]{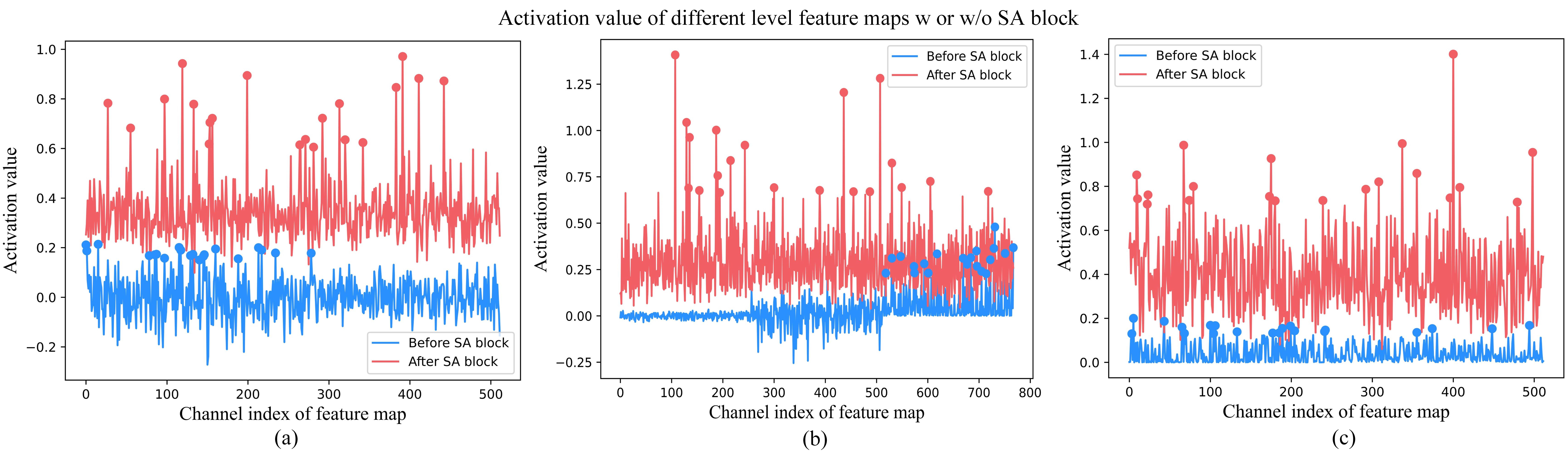}
\caption{Activation values of different levels of fine-grained feature maps w/ or w/o SA blocks. Activation values of low-level, middle-level and high-level fine-grained feature maps are plotted in sub-figures (a), (b) and (c), respectively.} 
\label{SA_block_all}
\end{figure*} \par 

\textbf{Baseline.} The high-level semantic features play an important role for identifying and locating large aircraft. In comparison to the detection results of RefineDet, precision rate of the Baseline increases to 0.829 and is 1.4\% higher than that of RefineDet at 0.5 IoU threshold. It may be because that less background information is accumulated at low-level feature layers through the top-down pathway, which reduces interference for aircraft detection to some extent. However, aircraft's high-level semantic features extracted by the Baseline is restricted resulting from the three-layer feature pyramid constructed by a truncated VGG-16 of the Baseline, which weaken its ability for identifying aircraft accurately. Therefore, distinct decreases emerge on $AP$, $AP^{.75}$ and $AP^{l}$ when detecting aircraft by Baseline, which are 1.0\%, 1.2\% and 1.3\% lower than those of RefineDet. \par 

\textbf{Effects of FF.} By merely propagating aircraft's textural details  forwardly, uninformative background interference may also be accumulated further, which leads to much confusion for aircraft discrimination. Specifically, score of $AP^{.75}$ achieved by Baseline (w/ FF) (the third column of Table~\ref{effect_modules}) decrease obviously and is 1.5\% lower than that of Baseline. \par  

\textbf{Effects of SA.} Aircraft's low-level textural details could be utilized effectively benefiting from powerful feature refinement ability of SA blocks, which promote the detection performance of our method greatly. In comparison to the detection results of Baseline (w/ FF) and Baseline (w/ FF+SA) (the third and the fourth columns of Table~\ref{effect_modules}), scores of precision, $F_1$, $AP$, $AP^{.75}$, $AP^{s}$, $AP^{m}$ and $AP^{l}$ achieved by Baseline (w/ FF+SA) are 4.4\%, 1.6\%, 1.4\%, 4.6\%, 5.6\% 1.6\% and 1.2\% higher than those of Baseline (w/ FF). It fully verifies effectiveness of SA blocks for refining aircraft's significant features and suppressing uninformative background. To inspect effects of SA blocks further, values of different fine-grained feature channels w/o or w/ SA blocks are also provided in Fig.~\ref{SA_block_all} in blue and red curves, respectively. Meanwhile, the top-k (k=20) corresponding feature channels sorted by activation values are also marked out with blue and red real dots. Obviously, peaks of feature channels after SA blocks (red points) distribute sparsely, however, densely and locally for peaks of initial feature channels, which illustrates that a wider range of aircraft's significant features across multi-level feature maps could be boosted. Especially for building the middle-level fine-grained feature map, the first 256 channel indexes at Fig.~\ref{SA_block_all} (b) are enhanced and a more balanced feature map could be obtained after enabling SA block, which proves powerful feature refinement ability of SA block.  

\textbf{Effects of DLCM.} Feature alignment by DLCM at lateral connections of the fine-grained feature pyramid promotes our method's ability for locating aircraft especially small aircraft further. Benefiting from powerful deformable modeling ability of DLCM, aircraft's high-level discrete features could be perceived accurately, leading to improvements on precision (P), $AP^{.75}$ and $AP^{l}$. However, scores of $AP^{s}$ encounters a sharp decrease and is 3.3\% lower than that of Baseline (w/ FF+SA). It might be because of insufficient semantic information and complex background interference exists at low-level feature maps, which restricts DLCM's ability for capturing aircraft's discrete and significant information.\par  

\textbf{Effects of ADM.} A tight correspondence between refined anchors and  features is constructed by Baseline (w/ FF+SA+DLCM+ADM), which significantly boosts performance of our method for detecting aircraft. Compared with detection results of Baseline (w/ FF+SA+DLCM) (the fifth row of Table~\ref{effect_modules}), of which the detection head is the same as that of RefineDet, remarkable improvements on all indicators are acquired by Baseline (w/ FF+SA+DLCM+ADM) (the last column of Table~\ref{effect_modules}). Specifically, scores of $AP$,$AP^{.75}$,$AP^{s}$ are 1.6\%, 2.5\% and 8.7\% superior to those of Baseline (w/ FF+SA+DLCM), respectively. \par
\begin{figure}[H]
\centering
\includegraphics[scale = 0.4]{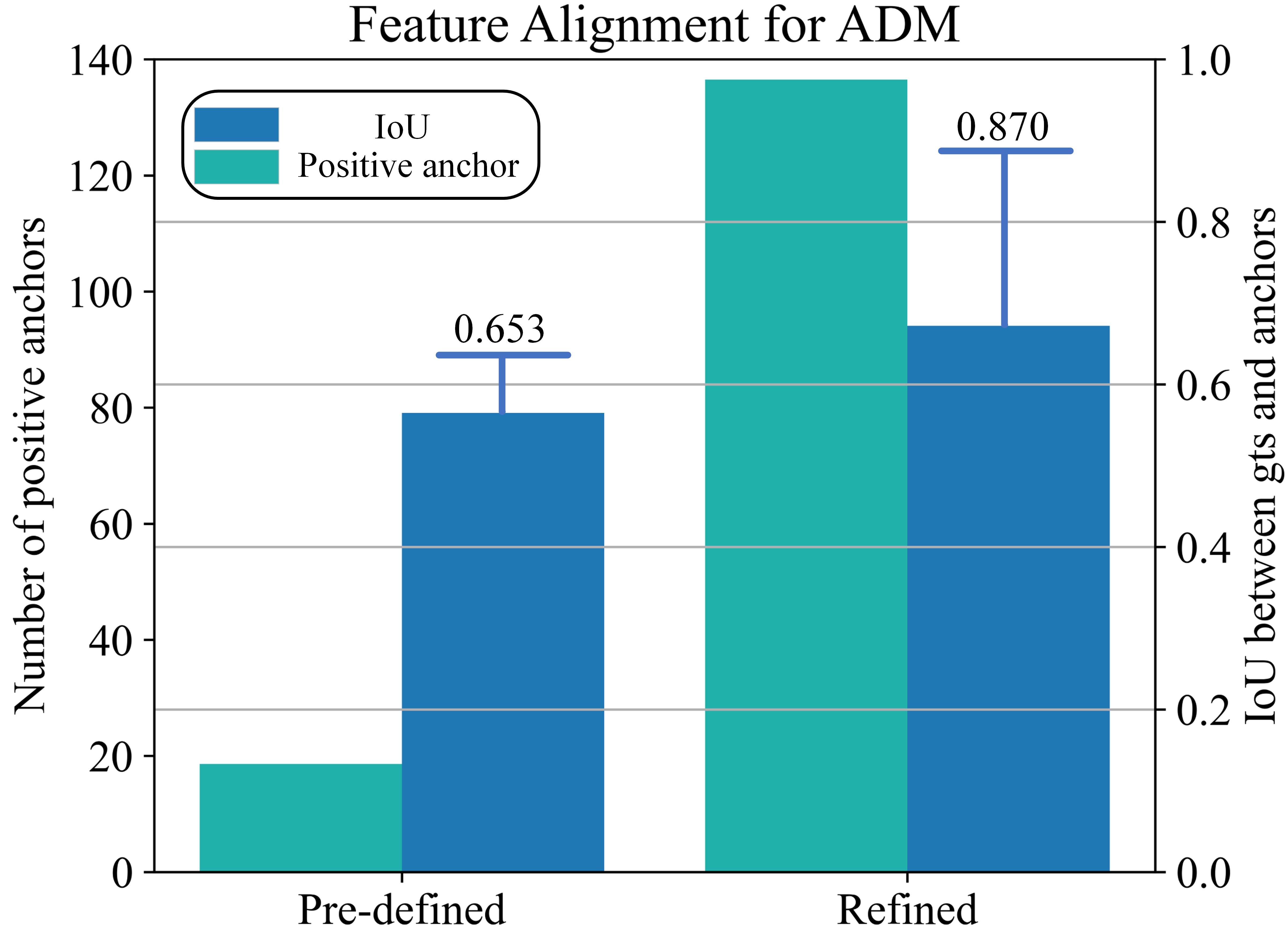}
\caption{Feature alignment of refined anchors for ADM. Blue and green bars show average IoU scores and positive anchor numbers achieved by initial and refined anchors at 0.5 IoU threshold, respectively. The horizontal lines indicate the maximal IoU scores acquired by initial and refined anchors.
\label{IoU_diff}} 
\end{figure}

 
Furthermore, number of positive anchors, values of average and maximal IoU between initial anchors and ground truths or refined anchors and ground truths are given by Fig.~\ref{IoU_diff}. Obviously, more positive anchors with higher IoU score are acquired by refined anchors than those of initial anchors. The average maximal IoU score between ground truths and refined anchors is 0.870 yet 0.653 for initial anchors, which further indicates positive guidance of refined anchors for ADM. \par 

 \par

\begin{figure*}
\centering
\includegraphics[scale = 0.41]{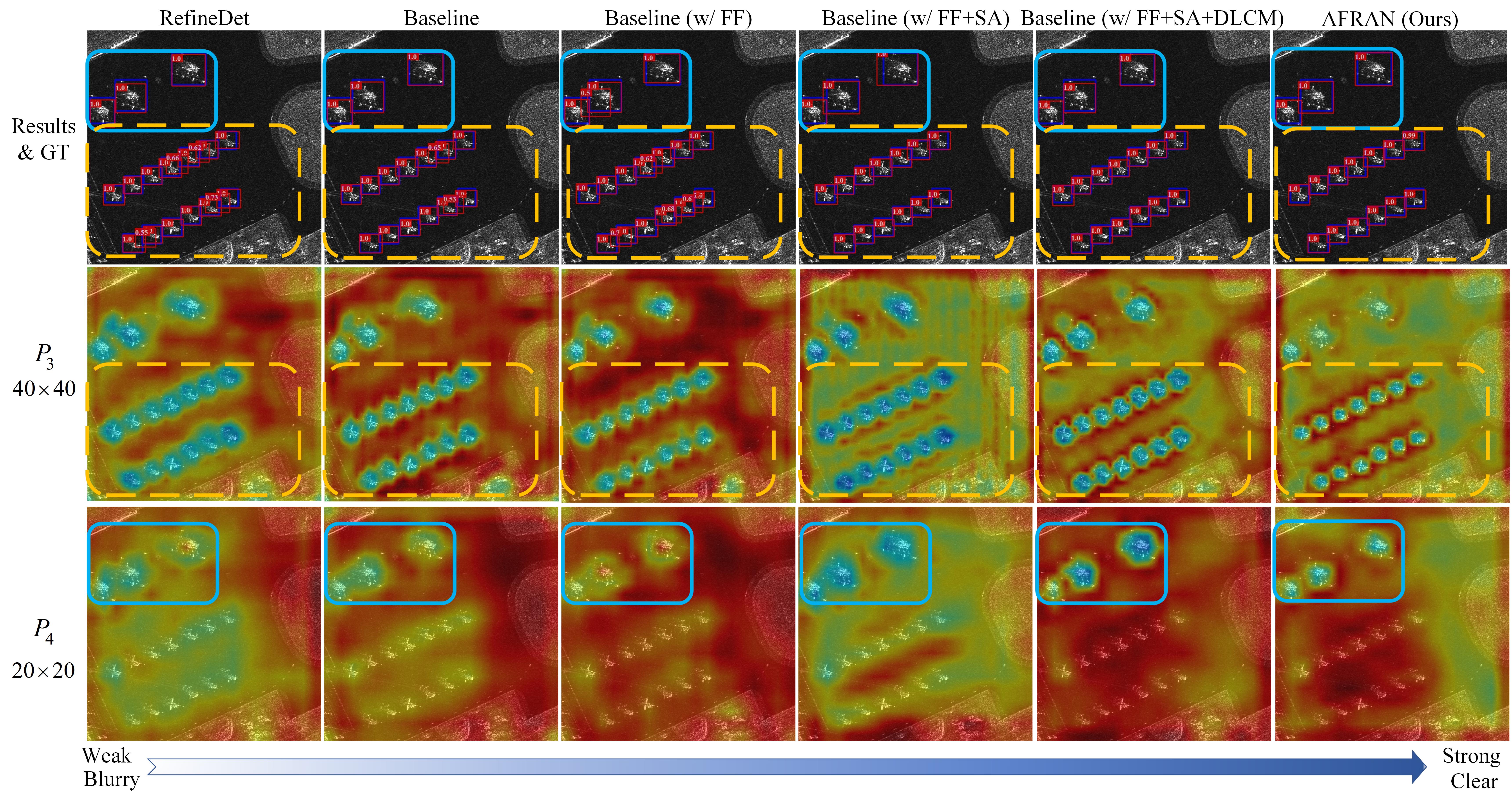}
\caption{Detection results and activation maps of specific fine-grained feature maps produced by our method equipped with the proposed sub-modules progressively.}
\label{inereff}
\end{figure*}\par 

\subsubsection{\textbf{\textit{Qualitative analysis}}}
To further investigate evolution of features maps acquired by the baseline equipped with the three sub-modules progressively, aircraft detection results and heat maps of specific fine-grained feature maps including $P_3$ and $P_4$, are provided in Fig.~\ref{inereff}. Some observations could be summed up as follows.\par 
\textbf{Clear boundaries between aircraft and background are obtained by Baseline (w/ FF).} Specifically, activation boundaries of small aircraft and surroundings at $P_3$ of Baseline (w/ FF) (the third column of Fig.~\ref{inereff}) are more distinct than those of the Baseline (the second column of Fig.~\ref{inereff}). Also, aircraft's activation areas are converged at $P_3$ of Baseline (w/ FF) yet divergent at identical feature areas generated by RefineDet (the first column of Fig.~\ref{inereff}). It might be because that fine-grained features could be enhanced at $P_3$ further by fusing aircraft's local textural details from $P_2$. Besides, less uninformative background information is propagated from topper feature maps, \ie $P_4$ in our method, which improves our method's ability for identifying aircraft and surroundings clearly.\par 
\textbf{Significant components of aircraft are specifically highlighted by Baseline (w/ FF+SA).} According to the fourth column of Fig.~\ref{inereff}, activation areas of small aircraft (dark blue areas encompassed by orange dotted rectangle) are distinct and the background are homogeneous (areas in green color) by enabling SA blocks. It might be because of the effective layer-aware feature refinement ability provided by SA blocks that consistency of background and specificity of aircraft at different feature maps could be maintained when fusing and refining low-level textural details and high-level semantic characteristics of aircraft. \par 
\textbf{Clear separations among densely arranged aircraft are obtained by Baseline (w/ FF+SA+DLCM).} According to the fifth column of Fig.~\ref{inereff}, separations among densely parked small aircraft are clearer than those at the same feature map $P_3$ of Baseline (w/ FF+SA). Besides, activation areas of large aircraft at $P_4$ are more centered than those extracted by Baseline (w/ FF+SA), which promotes our model to locate aircraft precisely (also proved in Table~\ref{effect_modules}). It might because that DLCM could capture irregular information of aircraft adaptively and avoid extracting much uninformative background interference by employing deformable convolutions instead of traditional convolutions equipped with axis-aligned kernels. \par 
\textbf{Hierarchy and consistency of aircraft's features at multi-scale feature maps are promoted by Baseline (w/ FF+SA+DLCM+ADM).} As illustrated by the last column of Fig.~\ref{inereff}, activation areas of small aircraft are more consistent than those at the same feature maps of Baseline (w/ FF+SA+DLCM). Also, aircraft with different scales could be represented hierarchically by different feature levels, which decreases aliasing across different feature maps. Specifically, activation areas of small and large aircraft are represented distinctly at $P_3$ and $P_4$, respectively. \par

\begin{table*}[!htbp]
\caption{Detection results of different CNN-based methods on test-set of aircraft sliced dataset.}
\centering 
\def\arraystretch{1.0} 
    \begin{tabular}{c|ccc|cccccc}
        \hline
        Models  &  P & R & $F_1$& $AP$ & $AP^{.5}$ & $ AP^{.75}$ & $ AP^{s}$ & $ AP^{m}$& $ AP^{l}$ \\
        \hline\hline
         \multicolumn{10}{c}{Two-stage methods} \\
         Faster R-CNN~\cite{ren2015faster} & 0.893 & 0.890 & 0.892  & 0.520 & 0.877 & 0.567 & 0.331 & 0.495 & 0.561 \\
         FPN~\cite{lin2017feature} &  0.870 & 0.909 & 0.889  &   0.530 & 0.886 &  0.598 & 0.366 & 0.509 & 0.562   \\
         Cascade R-CNN~\cite{cai2018cascade} & 0.898  & 0.889 & 0.893 &  0.539 & 0.876 & 0.594 & 0.398 & 0.517 & 0.577  \\
         DAPN~\cite{cui2019dense} & 0.885  & 0.830 & 0.857 &  0.448 & 0.851 & 0.413 & 0.250 & 0.420 & 0.489  \\
         \hline
         \multicolumn{10}{c}{One-stage methods} \\
         SSD~\cite{liu2016ssd} &  0.889  & 0.889 & 0.889 & 0.532 & 0.932 & 0.562 & 0.371 & 0.497 &  0.575  \\
         PADN~\cite{zhao2020attention} &  0.842  & 0.916 & 0.878 & 0.512 & 0.906 & 0.518 & 0.308 & 0.480 &  0.554  \\
         RefineDet~\cite{zhang2018single} & 0.815 & \textbf{0.935} & 0.871  &0.530 & 0.932 & 0.547 & 0.388 & 0.521 & 0.549  \\
         RPDet~\cite{yang2019reppoints} & 0.867  & 0.917 & 0.891  & 0.543 & 0.886 & \textbf{0.612} & 0.415 & 0.517 & \textbf{0.584} \\
         Ours (AFRAN)  & \textbf{0.904}  & 0.932 & \textbf{0.918} & \textbf{0.554} & \textbf{0.941} & 0.597 & \textbf{0.481} & \textbf{0.537} & 0.576   \\
        \hline
    \end{tabular}
    \label{CNN_based_accuracy}
\end{table*}

\begin{table*}[!htbp]
\centering
 \caption{\centering Running time and model complexity of different CNN-based methods}
    \renewcommand\tabcolsep{3.5pt} 
    \begin{tabular}{c|ccccc}
        \hline
        Models & Backbone & Input size & FPS & Params(M) & MAC(G) \\
        \hline\hline
         \multicolumn{6}{c}{Two-stage methods} \\
         Faster R-CNN~\cite{ren2015faster} & \multirow{3}{*}{ResNet-101}& \multirow{3}{*}{$\sim1333 \times 800$} & 6 & 51.75 &  921.07  \\
          FPN~\cite{lin2017feature} & & &  16 & 60.04 & 290.35    \\
          Cascade R-CNN~\cite{cai2018cascade} &  &  & 15  & 60.4 & 296.42  \\
          DAPN~\cite{cui2019dense} &  &  & 12  & 128.23 & 335.46  \\
         \hline \\
         \multicolumn{6}{c}{One-stage methods} \\
         SSD~\cite{liu2016ssd} & VGG-16 & $\sim512\times512$ & \textbf{90} & \textbf{25.22} & \textbf{88.84}  \\
         PADN~\cite{zhao2020attention} & ResNet-101 & $\sim640\times640$ & 33 & 102.16 & 247.12  \\
         RefineDet~\cite{zhang2018single} & VGG-16 & $\sim640\times640$  & 63 & 34.05 & 150.00 \\
         RPDet~\cite{yang2019reppoints} & ResNet-101 & $\sim1333 \times 800$ & 17  & 55.59 & 278.44  \\
         Ours (AFRAN) & VGG-16 & $\sim640\times640$ & 45 & 35.82 & 150.59  \\
        \hline
    \end{tabular}
     \label{time_space_cost}
\end{table*}
\subsection{Comparing with CNN-based methods}
\subsubsection{\textbf{\textit{Quantitative analysis}}}
The performance of our method are also evaluated by comparing with other CNN-based detectors on test set of the self-built aircraft sliced dataset and a large scene image. Moreover, two methods specially designed for object detection in SAR images including PADN~\cite{zhao2020attention} and DAPN~\cite{cui2019dense}, are also re-implemented and compared together.\par 
\textbf{Detection accuracy.} The detection accuracy of different methods on the test set of the self-built aircraft sliced dataset are provided in Table~\ref{CNN_based_accuracy}. Specifically, AP and $F_1$ achieved by our method are 1.1\% and 2.5\% higher than those of the second-place methods (RPDet, Cascade R-CNN), respectively. Although a slight superiority on $AP^{.75}$ and $AP^{l}$ is achieved by RPDet benefiting from its point-based aircraft representation strategy. However, its ability for identifying small aircraft is still unsatisfied. Besides, $AP^{.5}$ and $AP^{.s}$ scores achieved by RPDet are 5.5\% and 6.6\% lower than those achieved by our method, which are 0.941 and 0.481, respectively. It might be because of insufficient features of small aircraft extracted by the class-agnostic backbone. Obviously, competitive performance on metrics for evaluating location accuracy are achieved by most of the two-stage detectors benefiting from their advanced feature alignment strategies in comparison with classical one-stage detectors, \textit{e.g.,} SSD, RefineDet. However, our method could still locate aircraft competitively benefiting from its powerful feature alignment achieved by DLCM and ADM. Additionally, an obvious promotion is obtained by our method comparing with the two domain-specific methods including DAPN and PADN at all indicators. It might be because of their sub-optimal architectures and limited utilization for  aircraft's low-level details, which incurs a sharp degeneration for aircraft detection in SAR images.\par 
\textbf{Time and spatial complexities.} The detection speed and spatial complexities are evaluated and given by Table~\ref{time_space_cost}. Clearly, benefiting from truncated VGG-16 backbone as well as balanced three-layer feature pyramid, our method only takes up 35.82M and 150.59G on parameter volumes (Params) and multiply-accumulate operations (MAC), which is lower than those required by numerous two-stage detectors (Faster R-CNN, FPN, Cascade R-CNN, DAPN) and some one-stage detectors (RPDet, PADN). Meanwhile, our method achieves higher detection speed than RPDet and all two-stage detectors, which demonstrates its feasibility for real-time applications. However, our method runs slower than SSD and RefineDet causing by numerous parameters introduced by its complex sub-modules.

\subsubsection{\textbf{\textit{Qualitative analysis}}}
To inspect trends of detection performance, precision-recall curves of all compared CNN-based methods under various conditions of IoU thresholds and aircraft's sizes are given by Fig.~\ref{PR_CNN}. Undoubtedly, curve of our method (red curves in Fig.~\ref{PR_CNN} (a) to (f)) decreases much slightly than those of other methods with the increase of recall rate. Specifically,  distinct advantages of our method exist at the conditions of 0.5 and 0.5@0.05:0.95 IoU thresholds, \ie the Fig.~\ref{PR_CNN} (a) and (b). Besides, performance of our method is superior to other methods by a large margin for detecting small aircraft according to Fig.~\ref{PR_CNN} (d). It might be because of the fully exploration and refinement of aircraft's multi-level features in AFFM, which further illustrates the powerful discrimination ability for aircraft in SAR images achieved by our method. \par  

\begin{figure}[H]
  \centering 
  \includegraphics[scale = 0.25]{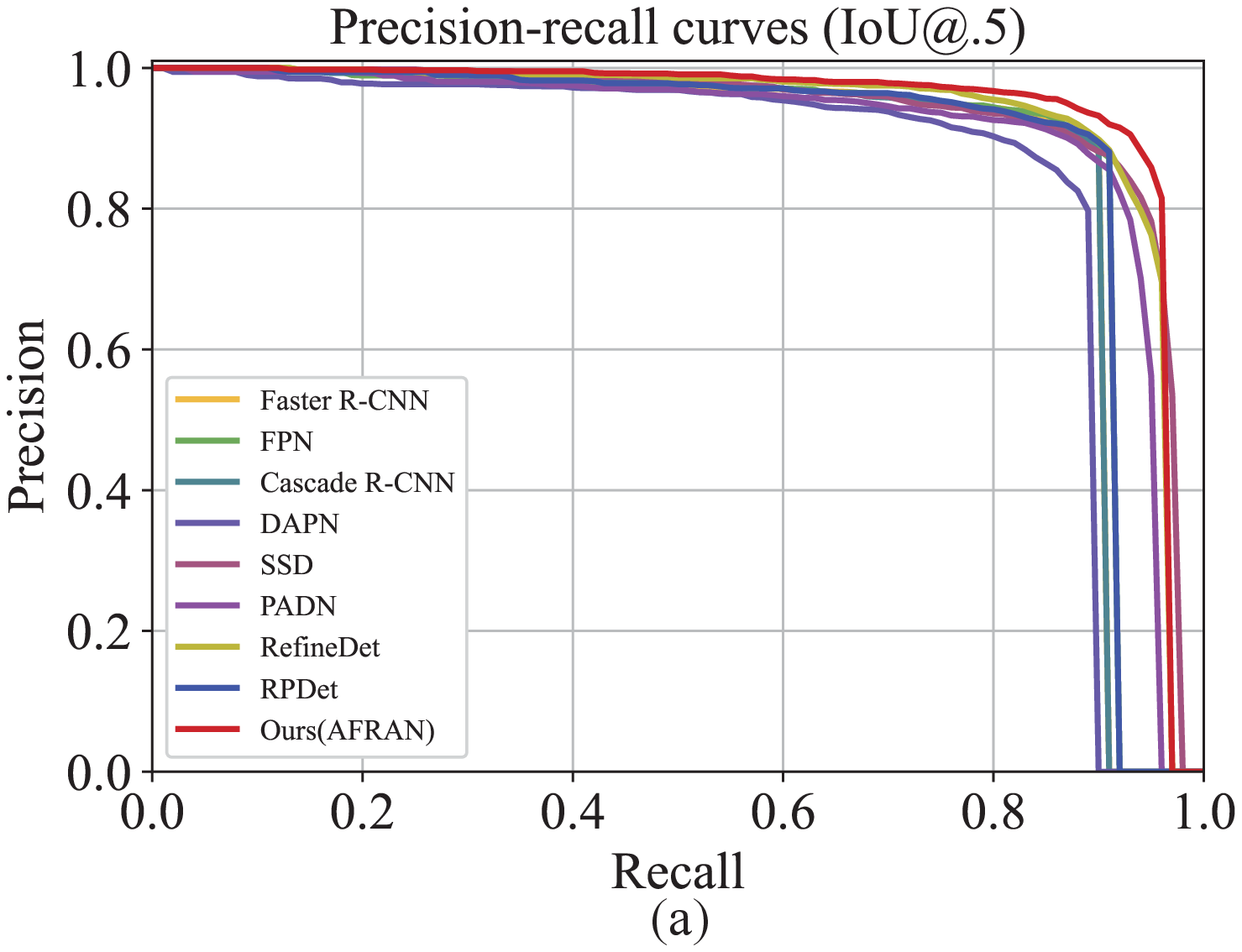}
  \includegraphics[scale = 0.25]{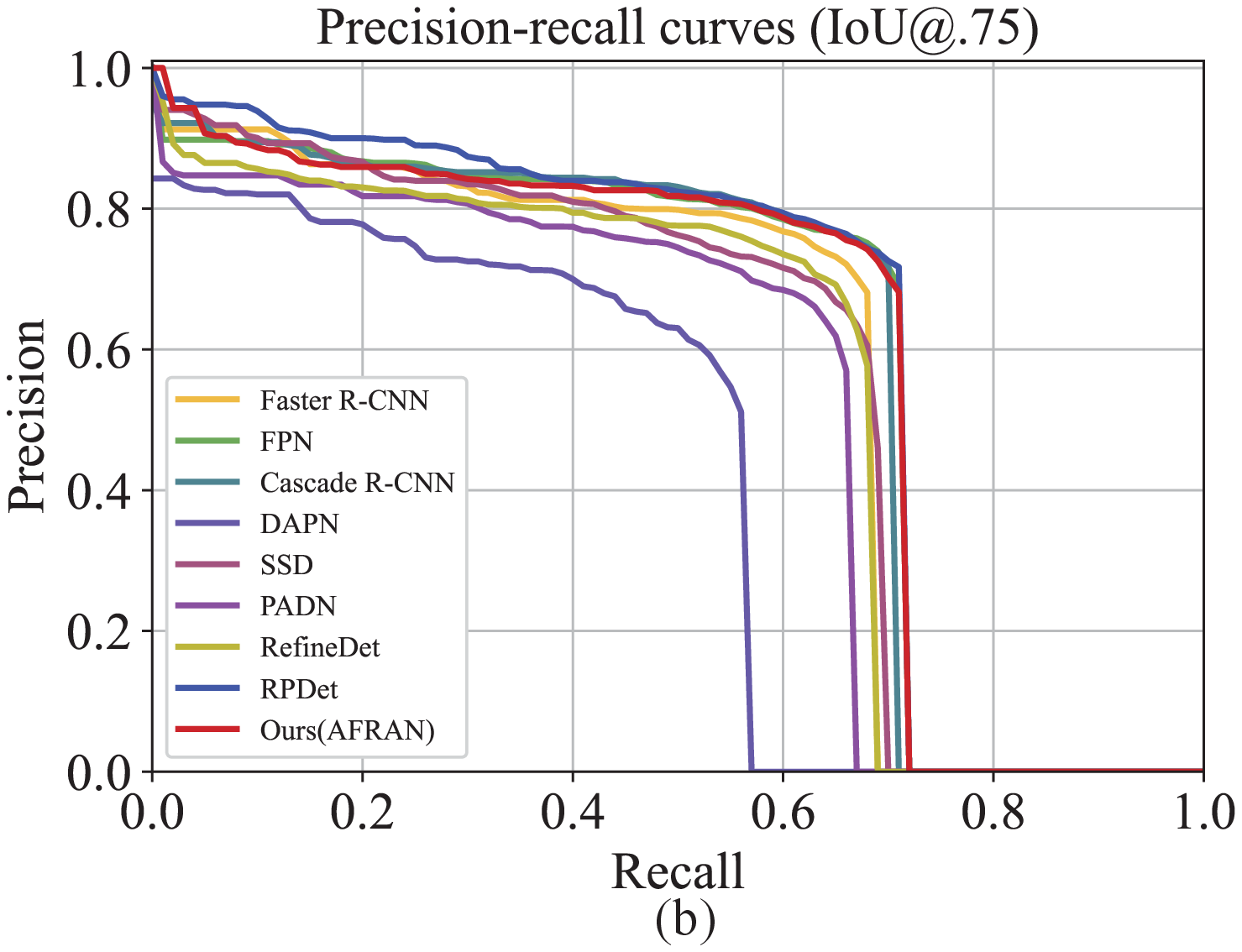}\\
  \includegraphics[scale = 0.25]{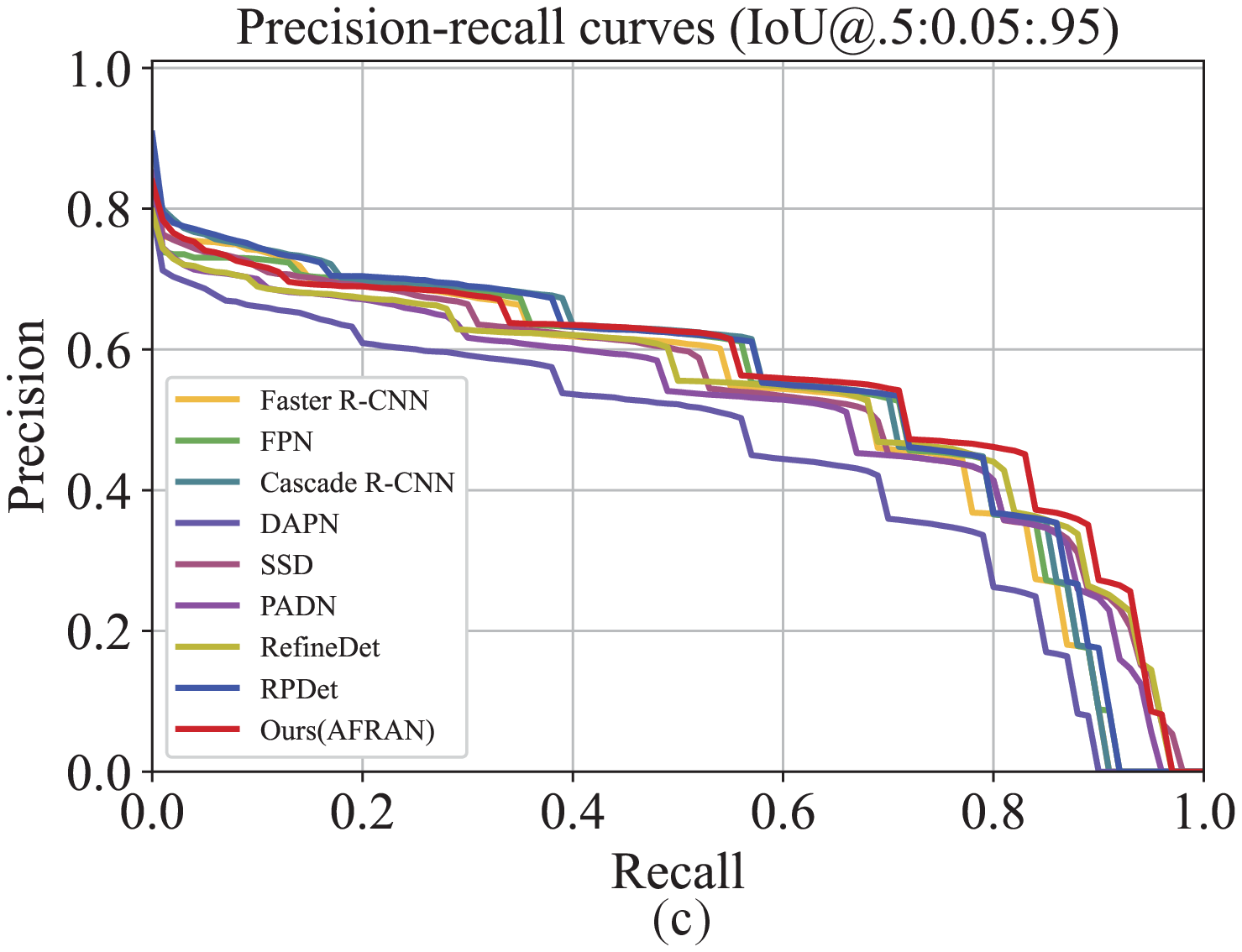}
  \includegraphics[scale = 0.25]{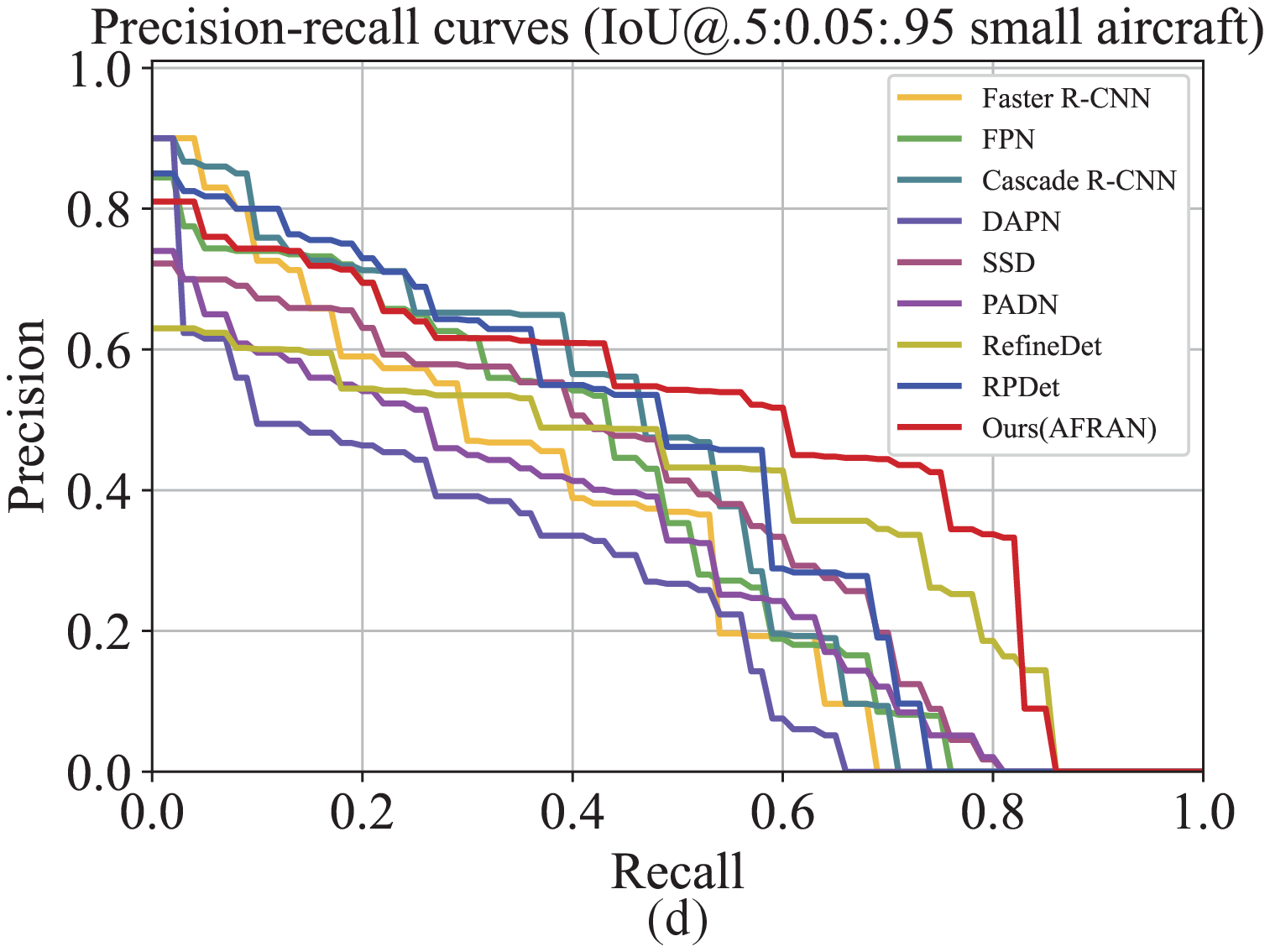}\\
  \caption{Continued.}
  \label{PR_CNN:1}
\end{figure}

\begin{figure}
  \ContinuedFloat 
  \centering 
  \includegraphics[scale = 0.25]{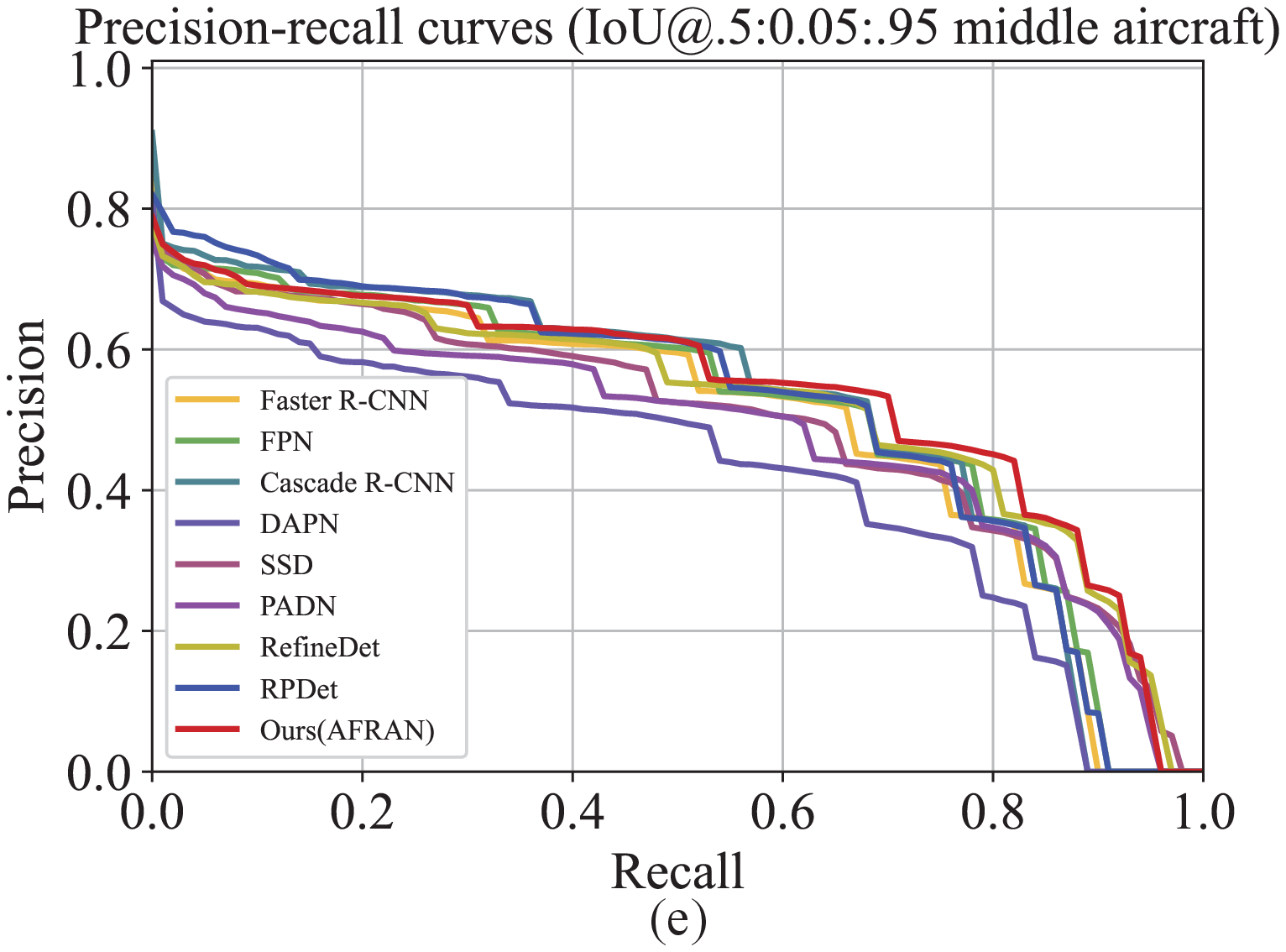}
  \includegraphics[scale = 0.25]{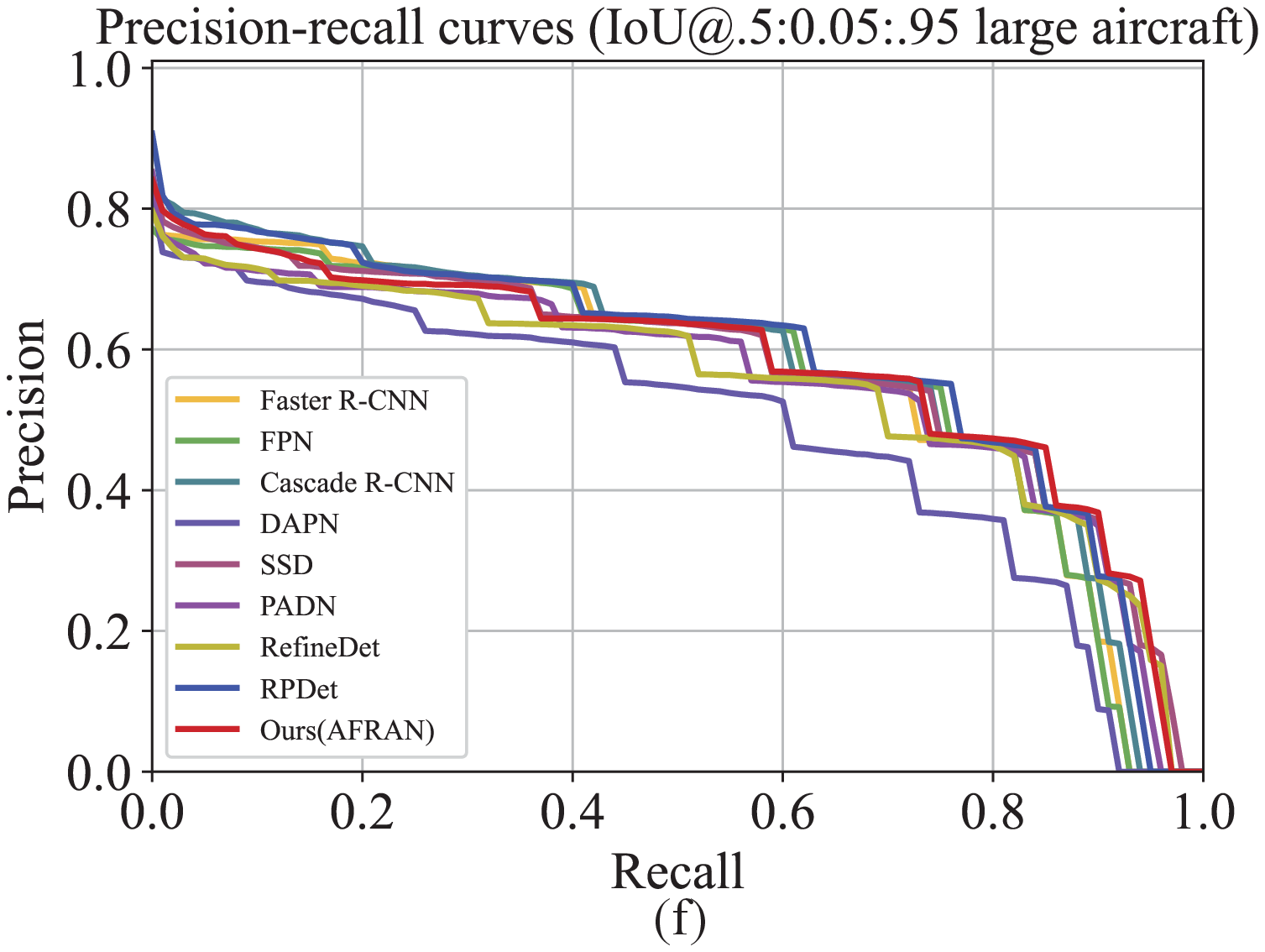}
  \caption{Precision-recall curves of the CNN-based methods for aircraft detection in SAR images.}
  \label{PR_CNN}
\end{figure}

\begin{figure}[H]
  \centering 
  \includegraphics[scale = 0.42]{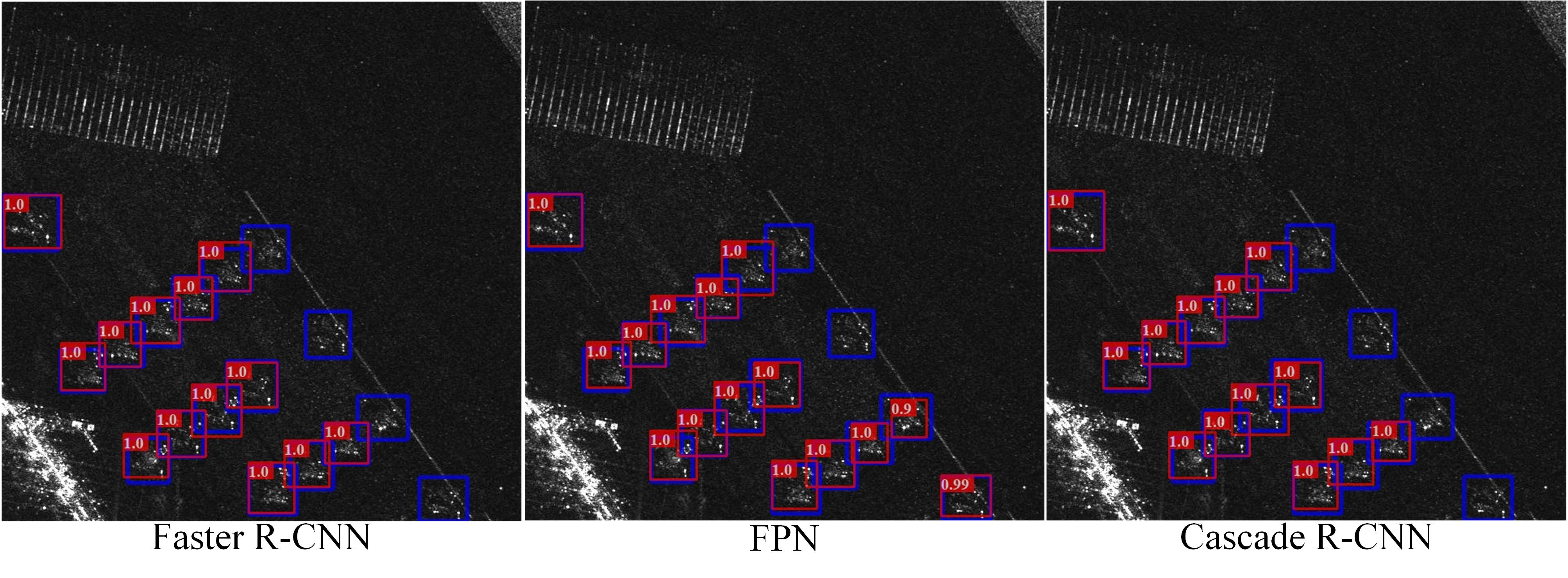}\\
  \includegraphics[scale = 0.42]{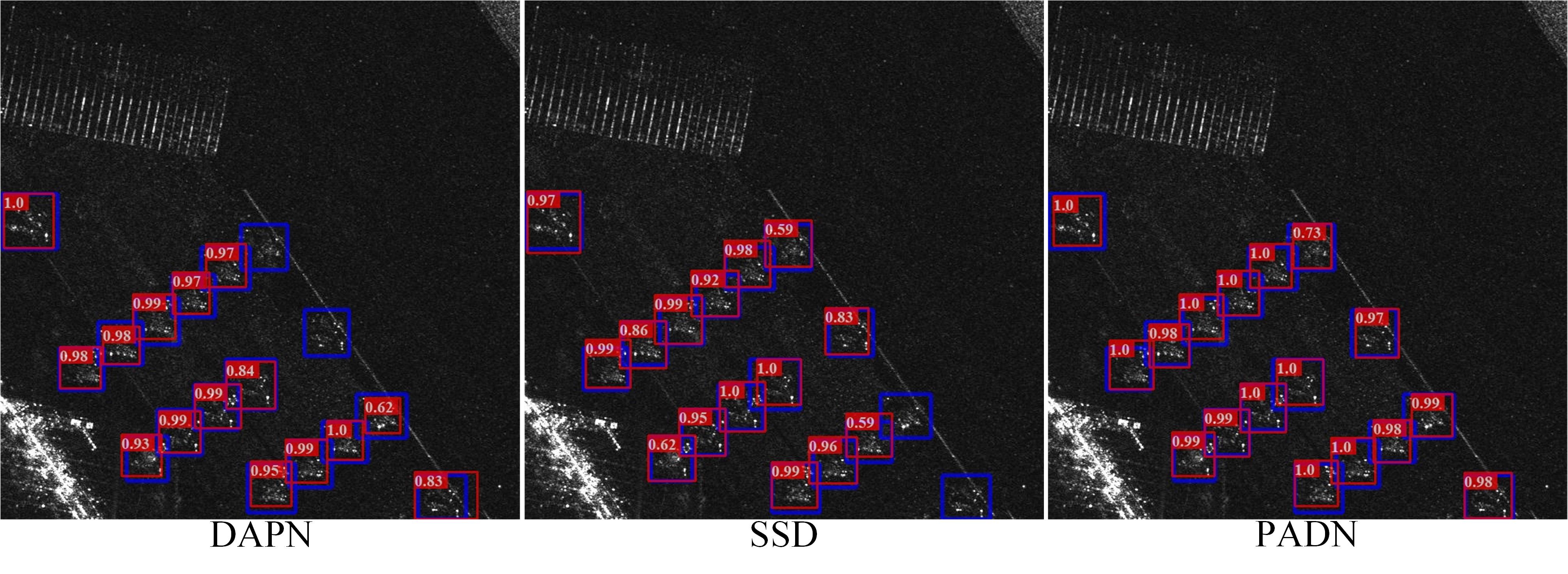}\\
  \includegraphics[scale = 0.42]{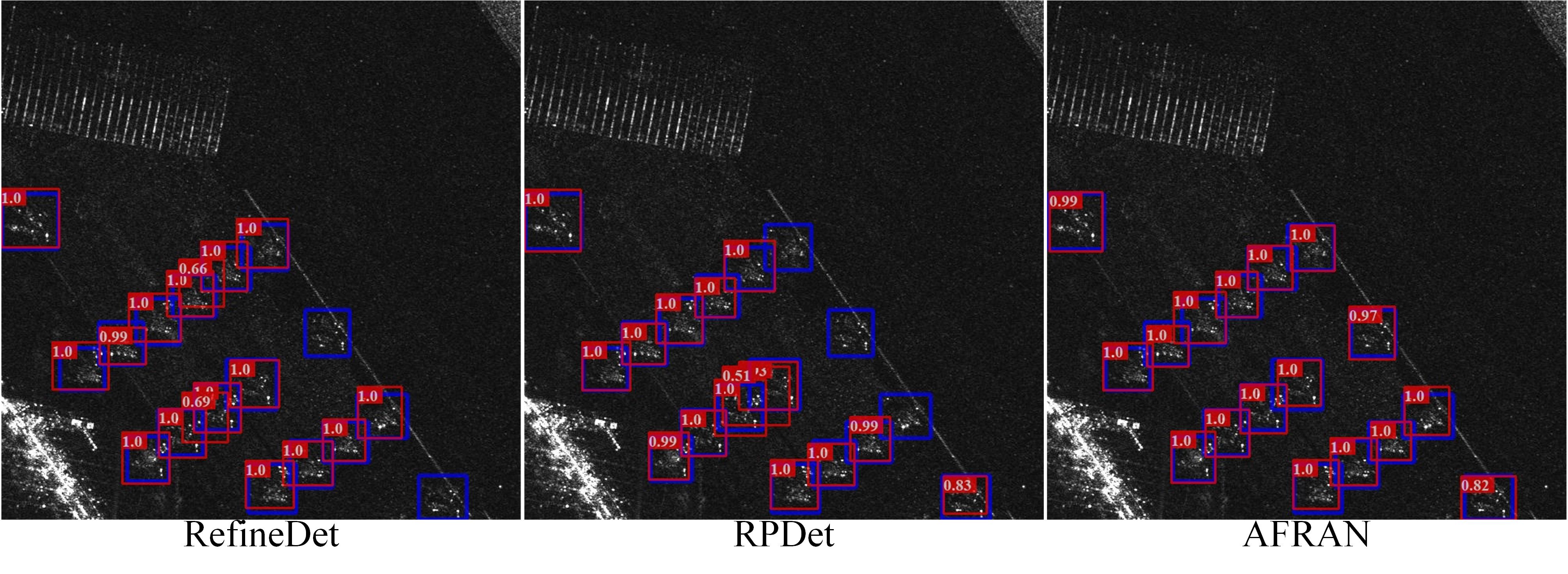}\\
  \includegraphics[scale = 0.42]{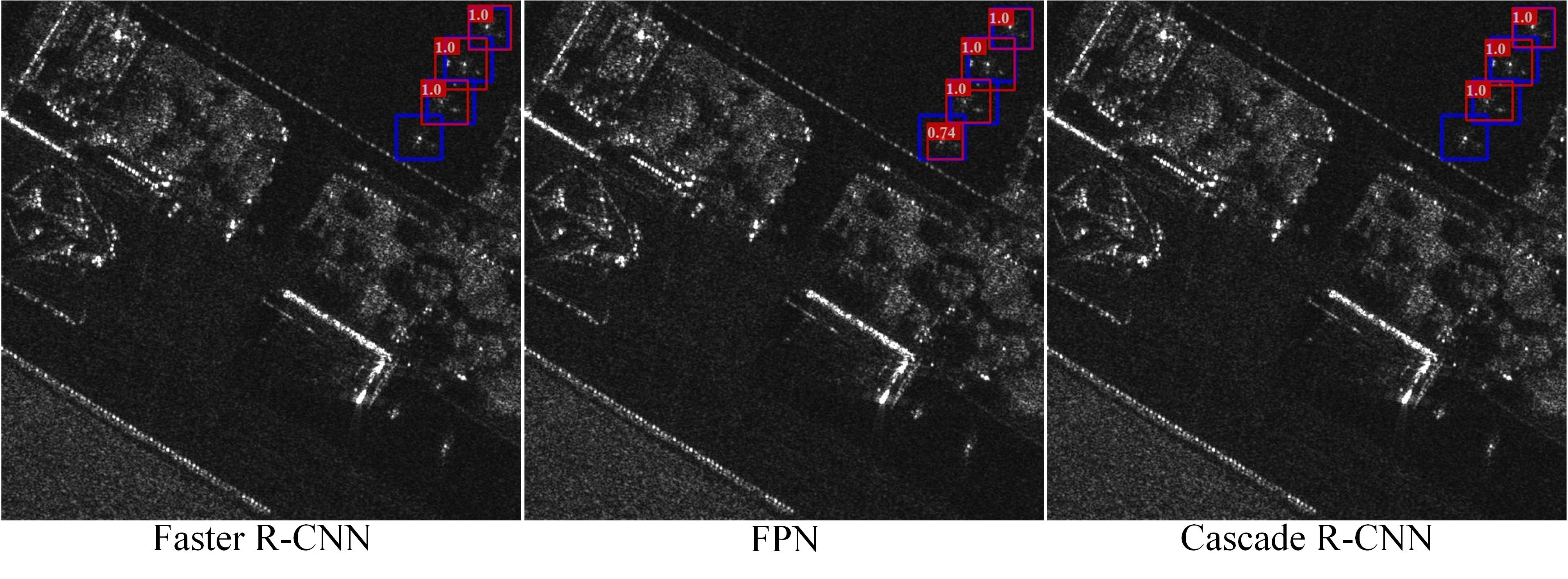}\\
  \includegraphics[scale = 0.42]{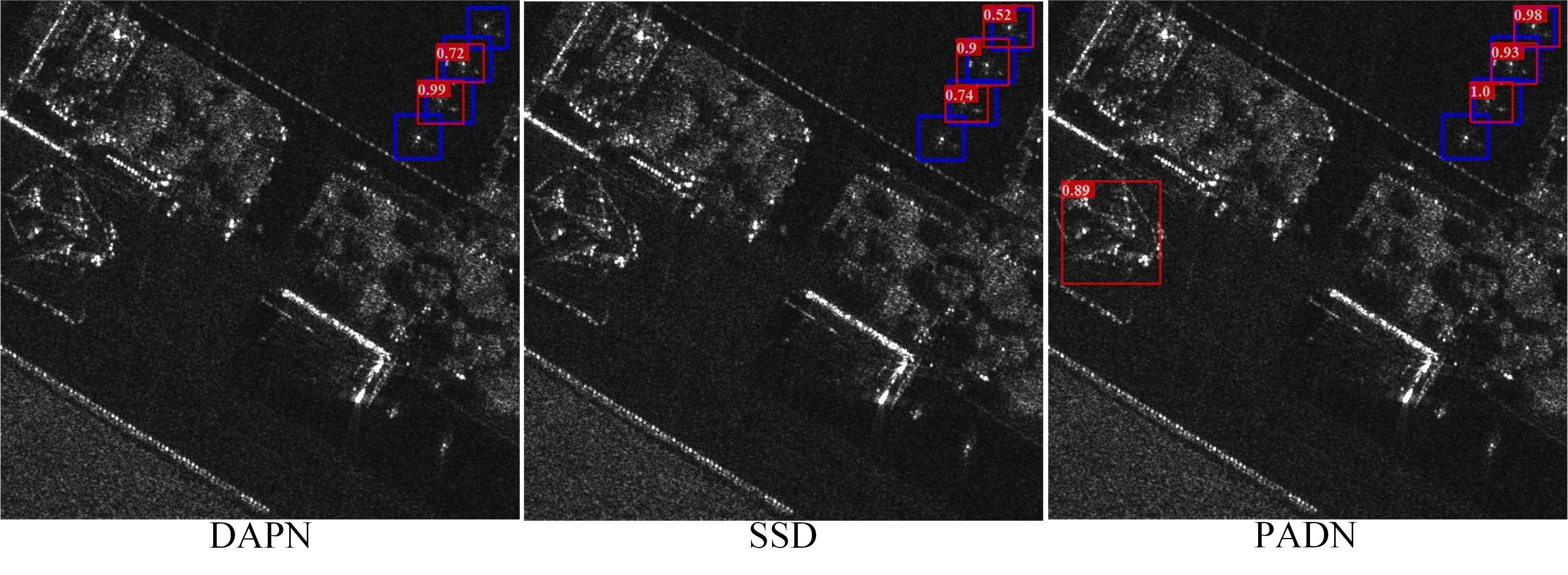}\\
 \includegraphics[scale = 0.42]{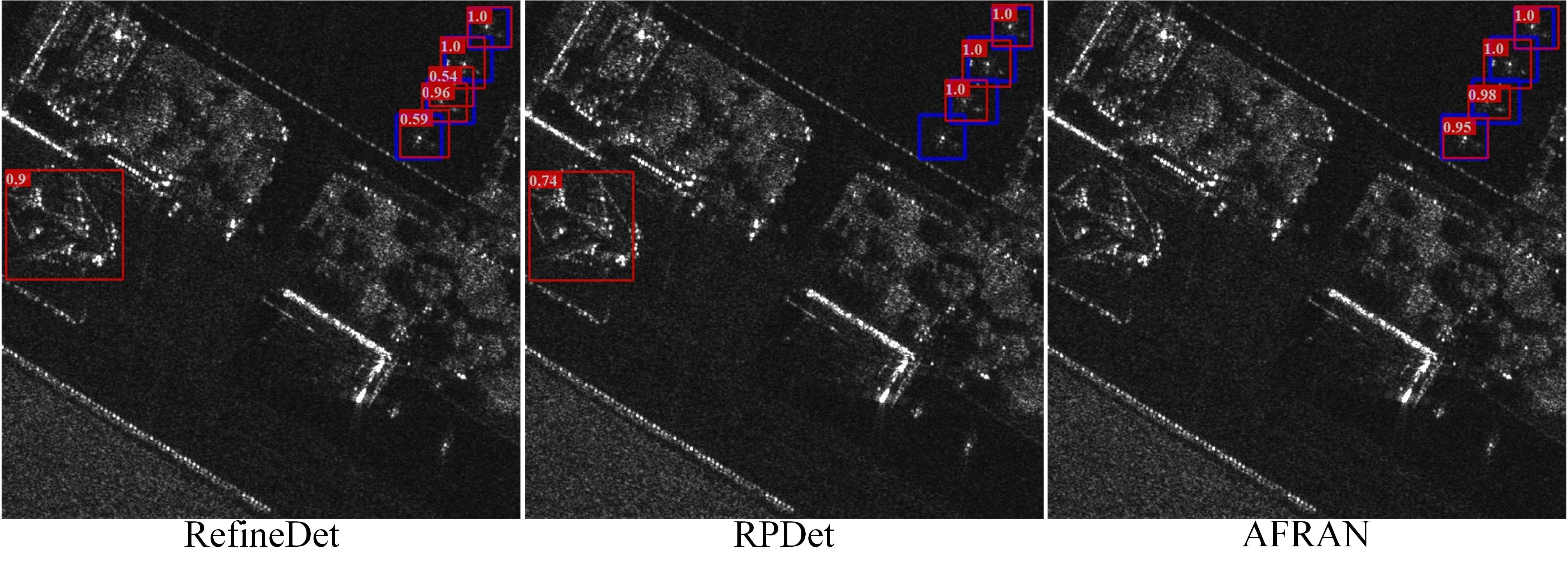}\\
  \caption{\textcolor{DarkBlue}{Continued}.}
  \label{visual_nudt_sadd:1}
\end{figure}

\begin{figure}[H]
\ContinuedFloat 
\centering
\includegraphics[scale = 0.43]{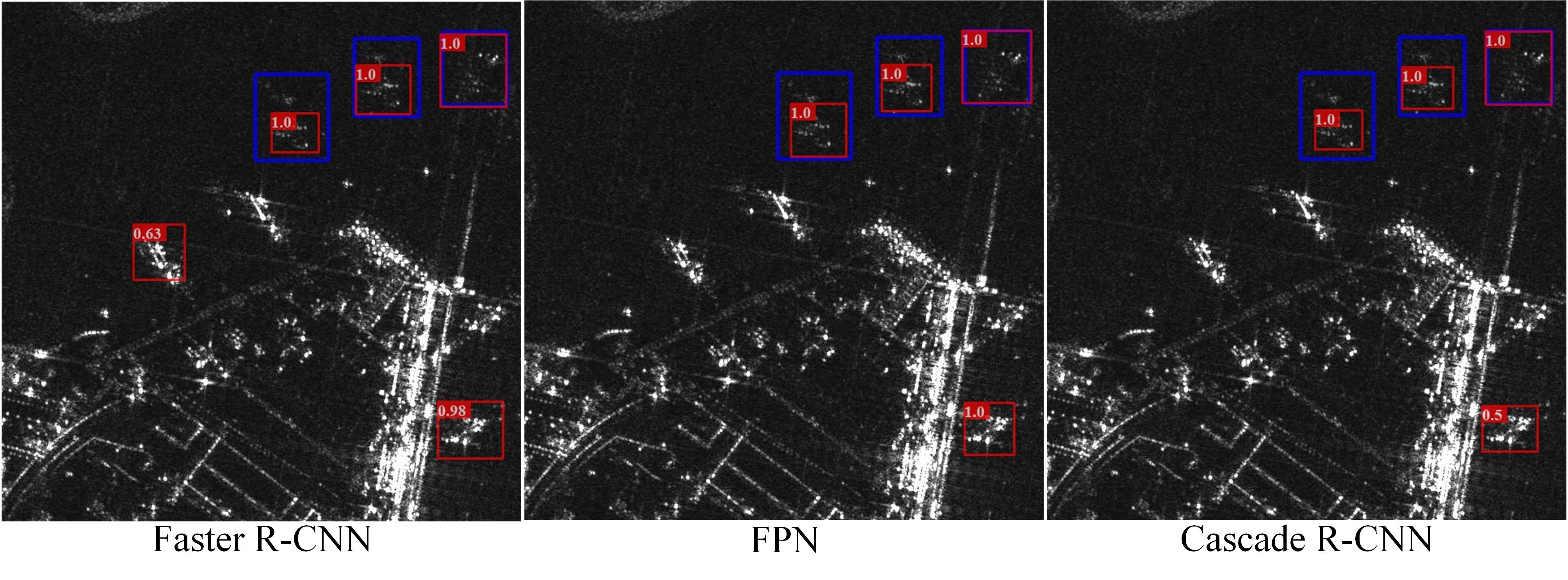}\\
\includegraphics[scale = 0.43]{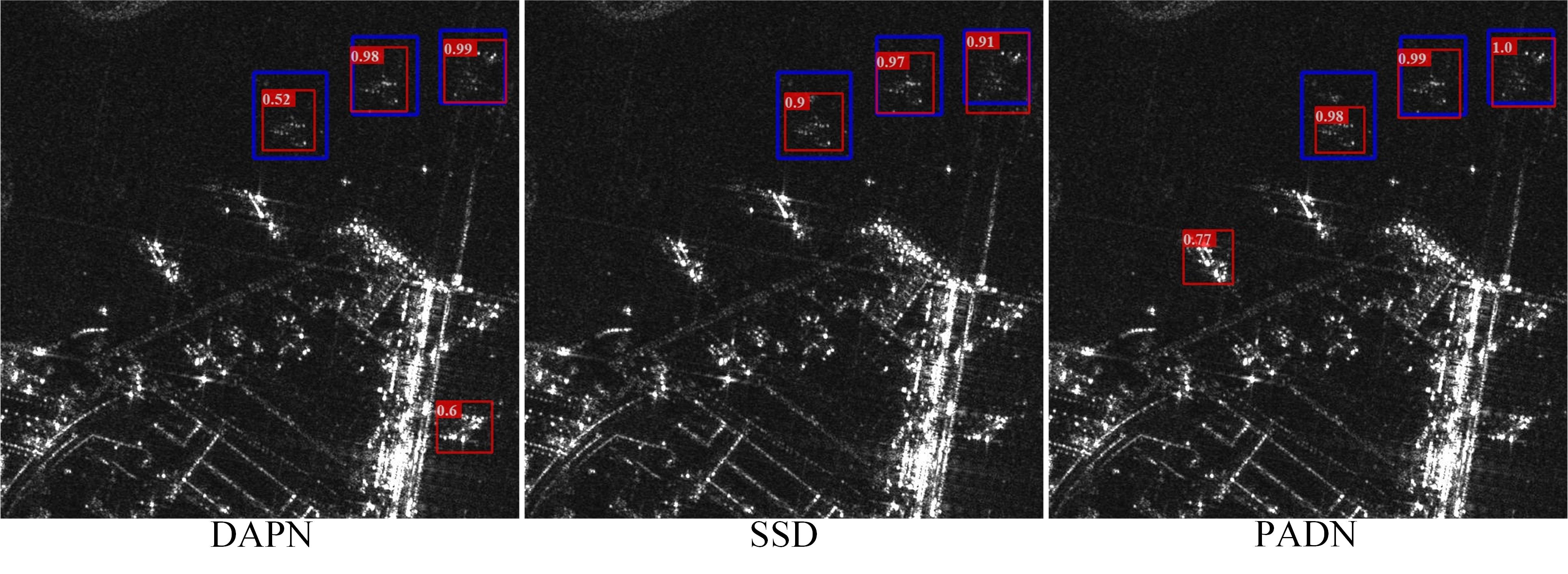}\\
\includegraphics[scale = 0.43]{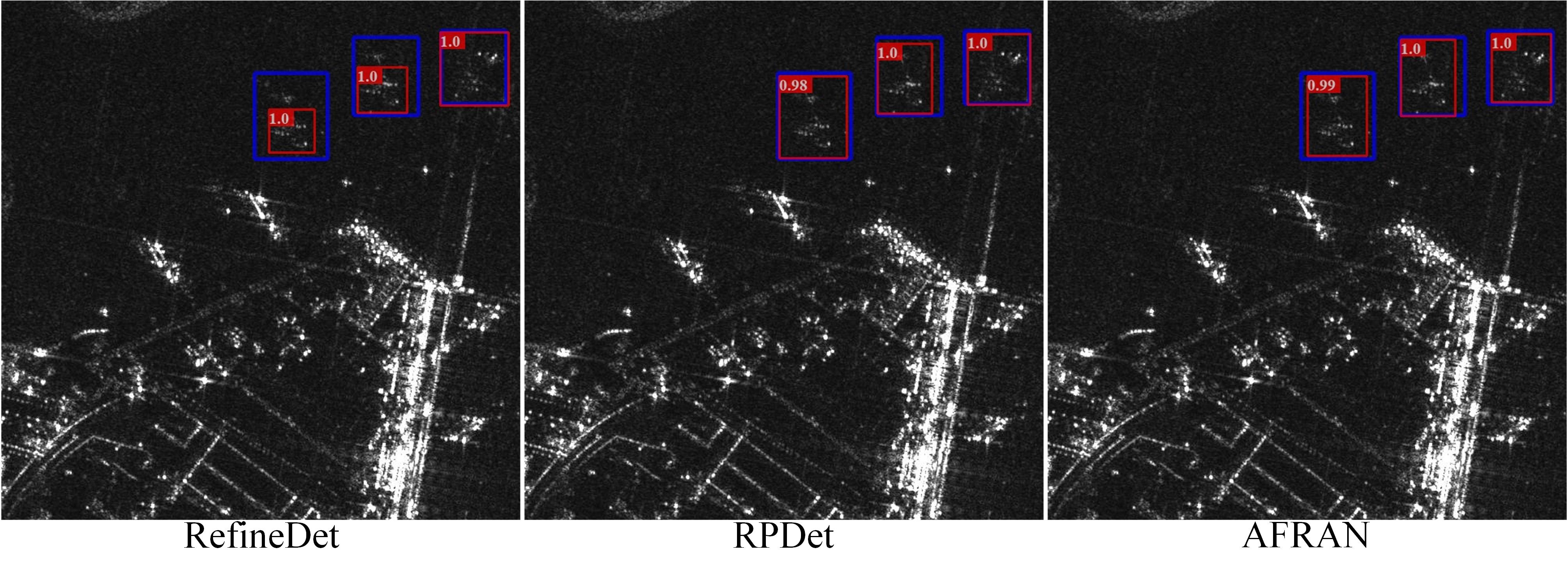}\\
\caption{\textcolor{DarkBlue}{Aircraft detection results of different CNN-based methods at three different situations}.}
\label{visual_nudt_sadd}
\end{figure} 

Additionally, we further inspect the visual detection performance of different CNN-based methods at three representative situations and the results are given by Fig.~\ref{visual_nudt_sadd}. In the first scene, aircraft with indistinct back-scattering information are densely parked together. In comparison with obvious false alarms detected by RefineDet and RPDet, and false negatives (the aircraft on the right side of the scene) missed by all methods except for PADN, our method detects these aircraft accurately with distinct separations. In the second scene, buildings on the left side of scene are detected as an aircraft by PADN, RefineDet, RPDet. Also, aircraft with poor back-scattering intensities (the first aircraft near the buildings) are missed by most compared methods. However, no false alarms are detected and false negatives are missed by our method. In the third scene, strong interference exists due to complex and angular lounge bridges as well as other ground facilities. In comparison to partial aircraft encompassed by other methods, \eg Cascade R-CNN, PADN and RefineDet, our method and RPDet all detect and encompass aircraft integrally benefiting from their flexible representation strategies for aircraft's discrete characteristics.\par
\begin{figure*}
\centering
\includegraphics[scale = 0.28]{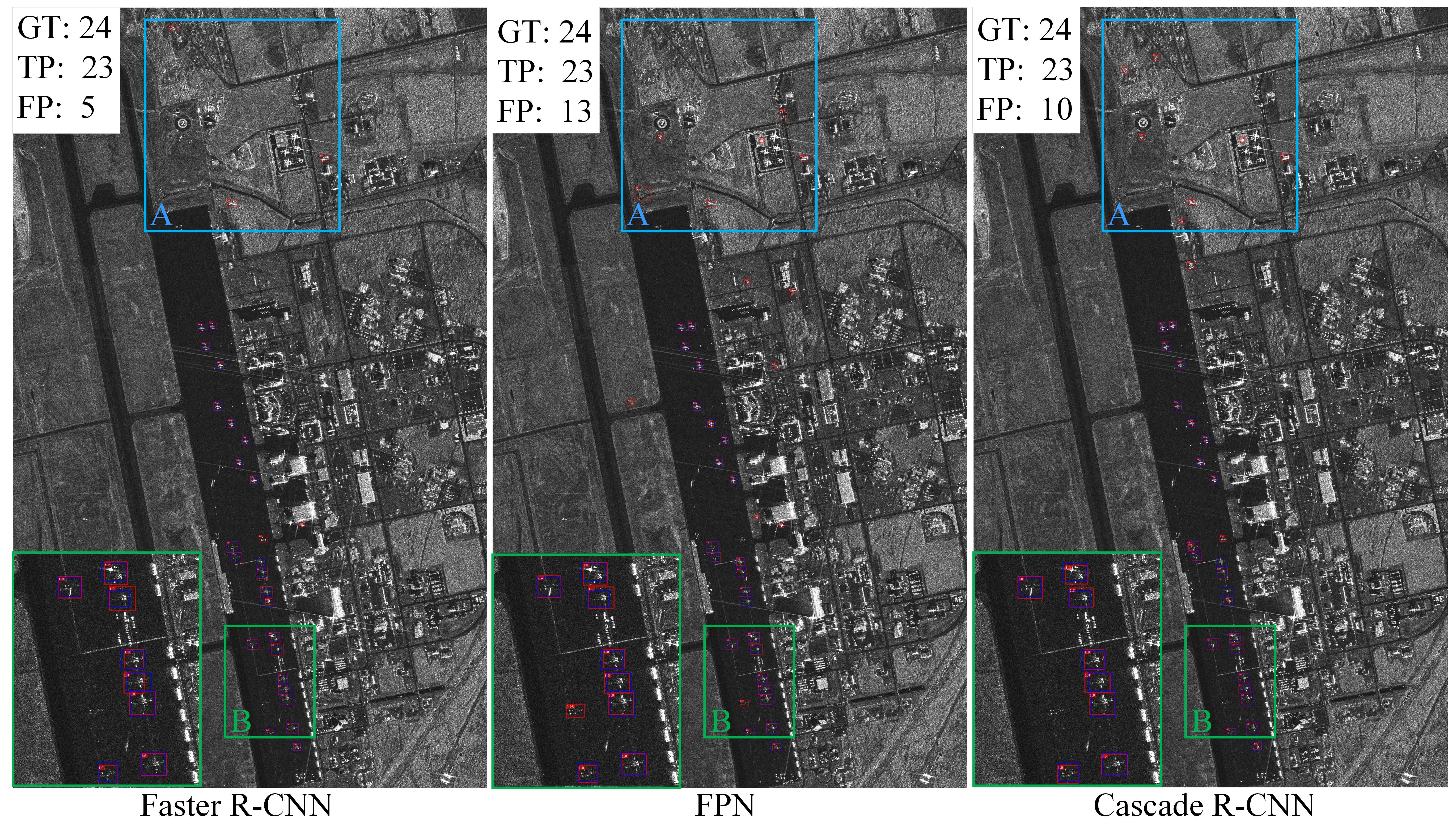}\\
\includegraphics[scale = 0.28]{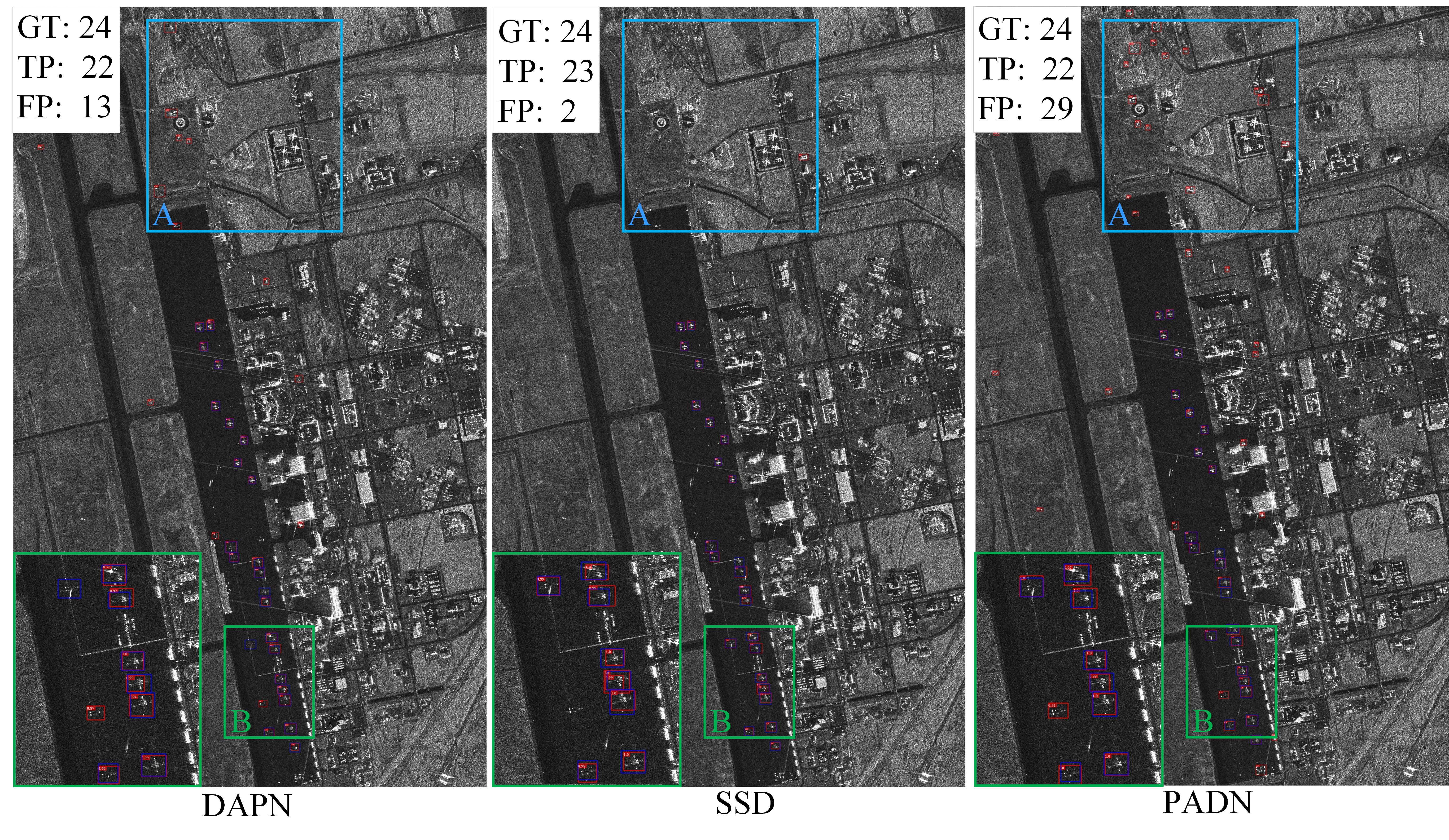}\\
\includegraphics[scale = 0.28]{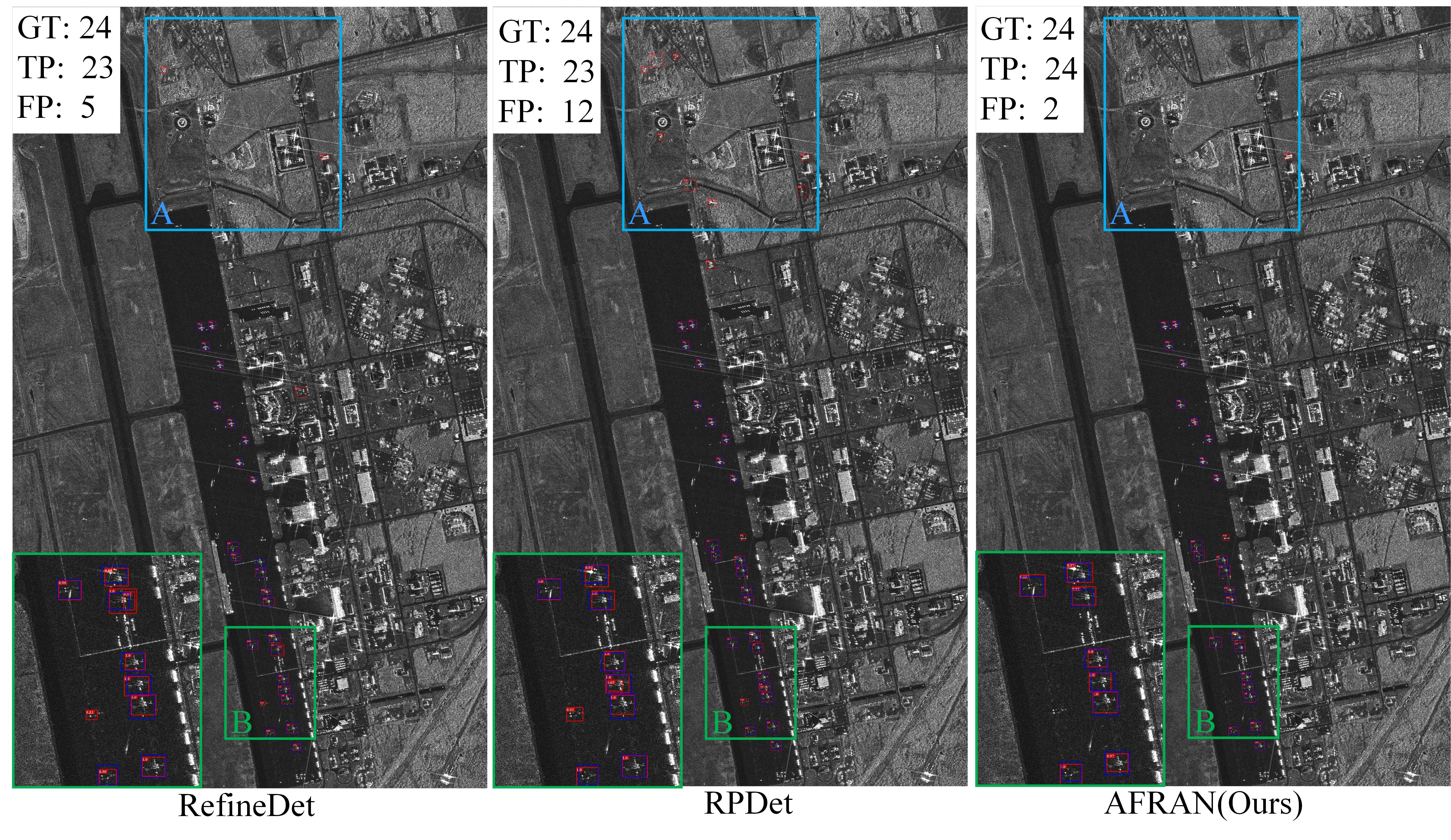}\\
\caption{Aircraft detection results over a large scene by different CNN-based methods.} 
\label{large_scene}
\end{figure*}

Apart from the above comparisons, the detection performance of different CNN-based methods over a large scene are investigated as shown in Fig.~\ref{large_scene}. \textcolor{DarkBlue}{To evaluate detection performance of different methods purely, only sliced cropping is adopted to satisfy input sizes of methods in advance. And the final detection results are acquired by mapping aircraft's coordination from partial slices to the large scene}. Obviously, our method could detect all aircraft with fewer false alarms than other compared methods. Most specifically, due to the lack of efficient feature refinement operations after feature aggregation in FPN, Cascade R-CNN, DAPN and PADN, surroundings within area A (the blue rectangle in Fig.~\ref{large_scene}) are easily recognized as aircraft. However, our method could discriminate aircraft and interference accurately. In terms of detecting aircraft with highly discrete back-scattering information, \eg aircraft parked at area B marked by the green rectangle in Fig.~\ref{large_scene}, aircraft's partial bodies are recognized as new ones by SSD, RefineDet and RPDet. Also, ground facilities within airport are detected as aircraft by FPN, DAPN and PADN, however, discriminated accurately by our method without any false alarms and negatives.\par 

\par 

\begin{table*}[t]
\caption{\label{Forward_fusion} Feature forward configuration of AFFM}
\centering 
\def\arraystretch{1.0} 
\begin{tabular}{cc|ccc|cccccc}
\hline
B  $\rightarrow$ M  & M $\rightarrow$ T & P & R & $F_1$  & $AP$ & $AP^{.5}$ & $ AP^{.75}$ & $ AP^{s}$ & $ AP^{m}$ & $ AP^{l}$  \\
\hline
\hline
\xmark & \xmark & 0.900  & 0.923 & 0.911  & 0.548 & 0.930 & \textbf{0.607} & 0.440 & 0.533 & 0.573 \\ 
\cmark & \xmark  & 0.902 & 0.931 & 0.917 & 0.547 & 0.939 & 0.602 & 0.462 & \textbf{0.538} & 0.561  \\
\xmark & \cmark & \textbf{0.963} & 0.762 & 0.851 & 0.499 & 0.856 & 0.532 & 0.380 & 0.488 & 0.519  \\
\cmark & \cmark & 0.904 & \textbf{0.932} & \textbf{0.918}  & \textbf{0.554} & \textbf{0.941} & 0.597 & \textbf{0.481} & 0.537 & \textbf{0.576}  \\
 \hline
\end{tabular} 
\end{table*}

\begin{table*}[t]
\caption{\label{SA_location} \textcolor{DarkBlue}{Locations and group configurations of SA blocks}}
\centering 
\def\arraystretch{1.0} 
\begin{tabular}{ccc|ccc|cccccc}
\hline
B  & M  & T & P & R & $F_1$ & $AP$ & $AP^{.5}$ & $ AP^{.75}$ & $ AP^{s}$ & $ AP^{m}$& $ AP^{l}$  \\
\hline
\hline
- & - & - & 0.883  & 0.932 & 0.907 & 0.540 & 0.939 & 0.579 & 0.405 & 0.520 & 0.568 \\ 
\cmark & - & - & 0.887 & 0.924 & 0.905 & 0.544 & 0.928 & 0.589 & 0.441 & 0.534 & 0.561  \\
- & \cmark  & - & 0.897 &  0.930 & 0.913 & 0.550 & 0.940 & 0.590 & 0.453 & 0.538 & 0.571  \\
- & - & \cmark & 0.899 & 0.929 & 0.914 & 0.548 & 0.939 & 0.576 & 0.427 & 0.532 & 0.576  \\
\cmark & \cmark & - & 0.893 & 0.922 & 0.907 & 0.550 & 0.932 & 0.602 & 0.422 & \textbf{0.541} & 0.568  \\
\cmark & -  & \cmark & 0.896 & 0.930 & 0.913 & 0.542 & 0.931 & 0.569 & 0.431 & 0.531 & 0.566 \\
- & \cmark & \cmark & 0.893 & 0.922 & 0.907 & 0.553 & 0.934 & 0.601 & 0.418 & 0.539 & 0.577 \\
\cmark(1) & \cmark(1) & \cmark(1) & \textbf{0.911} & 0.925 & 0.918 &  0.547 & 0.924 & \textbf{0.606} & 0.446 & 0.536 & 0.574  \\
\cmark(2) & \cmark(3) & \cmark(2) & 0.904 & \textbf{0.932} & \textbf{0.918} &  \textbf{0.554} & \textbf{0.941} & 0.597 & \textbf{0.481} & 0.537 & \textbf{0.576}  \\

 \hline
\end{tabular} 
\end{table*}

\begin{table*}[t]
\caption{\label{DLCM_params} Different configurations of dilation rates of DLCM}
\centering 
\def\arraystretch{1.0} 
\begin{tabular}{c|ccc|cccccc}
\hline
B$\rightarrow$ M $\rightarrow$ T  & P & R & $F_1$ & $AP$ & $AP^{.5}$ & $ AP^{.75}$ & $ AP^{s}$ & $ AP^{m}$& $ AP^{l}$\\
\hline
\hline
1,1,1  & \textbf{0.904}  & \textbf{0.932} & \textbf{0.918} & \textbf{0.554} & \textbf{0.941} & \textbf{0.597} & \textbf{0.481} & \textbf{0.537} & 0.576 \\ 
3,2,1  & 0.894 & 0.931 & 0.912 & 0.551 & 0.933 & 0.593 & 0.454 & 0.533 & 0.576  \\
6,4,2  & 0.892 &  0.930 & 0.911 & 0.550 & 0.941 & 0.596 & 0.427 & 0.528 & \textbf{0.578}  \\
 \hline
\end{tabular} 
\end{table*}

\subsection{More Discussion}
Diversified parameter configurations also affect the performance of our method, which are discussed in this subsection. \par 
 \textbf{Feature forward of AFFM.} In our method, two feature forward propagation pathways plotted as orange arrows in Fig.~\ref{AFRAN}, are specially designed for enhancing representation ability of the fine-grained middle $P_3$ and the top $P_4$ feature maps for aircraft's low-level details. Table~\ref{Forward_fusion} illustrates the detection results of our method configured with different feature forward pathways in AFFM. For a fair reference, the detection results of AFRAN without any feature forward branches in AFFM acting as a baseline, are given by the first row of Table~\ref{Forward_fusion}. Obviously, a 2.2\% promotion on $AP^{s}$ is achieved after combining low-level details of aircraft at Conv4\_3 with semantic features existing at the middle layer Conv5\_3 base on the baseline. However, a sharp decrease emerges on all indicators when merely introducing middle-level features to the topmost feature layer Conv7 due to the limited supervision of low-level details. When equipped with the two feature forward branches, a competitive performance, \eg $AP^{.5}$ is 4.1\% higher than that of the baseline, is acquired by our method benefiting from abundantly detailed and semantic features represented by the multi-scale fine-grained feature pyramid.\par 
\textbf{Locations and groups of SA blocks.} Due to various and complex semantic information at multi-scale feature maps, how the SA blocks placed and the number of SA blocks are also worth exploration. Table~\ref{SA_location} shows the contributions of SA blocks located at different feature layers. Most specifically, enabling SA blocks at a single feature layer, \ie the first four rows of Table~\ref{SA_location}, could boost the detection performance of our method, \eg $AP$ and $F_1$, resulting from powerful feature refinement donated by their well-designed layer-aware attentions. Especially, an obvious improvement is achieved after enabling SA blocks at the middle feature map when building $P_3$. It may be because of fully consideration of aircraft's low-level and high-level features remaining at the three feature maps. \textcolor{DarkBlue}{Besides, as can be seen from the fifth to the seventh columns of Table~\ref{SA_location}, most AP metrics increase when enabling SA blocks at two feature layers benefiting from effective utilization of complementary semantic features at different feature layers}. Eventually, by applying SA blocks at all feature levels, $AP$, $AP^{0.75}, AP^{s}$ achieved by our method, are 1.4\%,1.8\% and 7.6\% higher than those of the baseline, which proves superior characteristics of SA blocks for feature refinement. Additionally, the effects of the number of attention groups are shown at the eighth and ninth rows of Table~\ref{SA_location}. Due to a unspecific attention mechanism provided by the unified weights when setting attention groups of SA blocks to 1, a slight decrease on $AP$ appears leashing their ability for refining semantic information at specific feature levels.\par 
\textbf{Receptive fields of DLCM.} \textcolor{DarkBlue}{Capturing discrete features of aircraft accurately calls for a sophisticated design of convolutional receptive fields.} Table~\ref{DLCM_params} shows the detection results of our method configured with various dilation rates in DLCM. The larger the dilation rates are, the wider the receptive fields captured by DLCM. It could be observed that most of specific and comprehensive metrics, \eg precision $(P)$, $AP^{s}$, $AP^{m}$, $F_1$, $AP$, obtained by our method decreases continually with the increase of dilation rates from bottom to top feature layers. It might be because of much uninformative background captured by enlarged receptive fields, which causes to serious interference on feature extraction for small aircraft. Therefore, setting all dilation rates of deformable convolution within DLCM to 1 is a moderate trade-off for extracting discrete characteristics of aircraft. 

\section{Conclusion}
In this paper, an Attentional Feature Refinement and Alignment Network (AFRAN) is proposed for aircraft detection in SAR images by carefully taking aircraft's domain knowledge and challenges into consideration. Three significant components including Attention Feature Fusion Module (AFFM), Deformable Lateral Connection Module (DLCM) and Anchor-guide Detection Module (ADM), were introduced for adaptively refining and aligning significant discrete features of aircraft. Ablation studies conducted on a self-built aircraft sliced dataset verified the powerful feature aggregation and refinement abilities of AFFM and effective feature alignment ability of DLCM and ADM by adopting them at the lateral connections and the detection heads of a three-layer fine-grained feature pyramid, respectively. Extensive experiments conducted on test set of the aircraft sliced dataset and a large scene SAR image demonstrated the topmost detection accuracy as well as competitive speed and spatial complexities achieved by our method in comparison with other domain-specific, \eg DAPN, PADN, and general CNN-based methods, \eg Faster R-CNN, FPN, Cascade R-CNN, RefineDet and RPDet. In the future, a further combination between characteristics of aircraft and network design will be investigated to make great promotions for aircraft detection in SAR images. 

\bibliographystyle{IEEEtran}
\bibliography{bib.bib}

\end{document}